\documentclass[10pt,journal,compsoc]{IEEEtran}
\ifCLASSOPTIONcompsoc
  \usepackage[nocompress]{cite}
\else
  \usepackage{cite}
\fi

\ifCLASSINFOpdf
\else
\fi

\usepackage{amsmath}
\usepackage{algorithmic}
\usepackage{graphicx}
\usepackage{color}
\usepackage{amssymb}
\usepackage{mathrsfs}
\usepackage{multirow}
\usepackage{array}
\usepackage{url}
\usepackage{wrapfig}
\usepackage{xspace}
\usepackage{booktabs}

\usepackage{bm}
\usepackage{siunitx}
\usepackage[super]{nth}

\usepackage[pagebackref=true,breaklinks=true,letterpaper=true,colorlinks,linkcolor=blue,citecolor=blue,bookmarks=false]{hyperref}

\usepackage{bbm}
\usepackage{colortbl}
\usepackage{lipsum}

\makeatletter
\DeclareRobustCommand\onedot{\futurelet\@let@token\@onedot}
\def\@onedot{\ifx\@let@token.\else.\null\fi\xspace}

\def\eg{\emph{e.g}\onedot} 
\def\ie{\emph{i.e}\onedot} 
 
 \def\vs{\emph{vs}\onedot}
 
\def\etal{\emph{et al}\onedot}
\makeatother

\makeatletter

\newcommand{\Rmnum}[1]{\expandafter\@slowromancap\romannumeral #1@}
\makeatother

\newcommand{\netname}{RSC-Net}
\newcommand{\revise}[1]{#1}
\newcommand{\revises}{}

\begin{document}
\title{\revise{3D Human Pose, Shape and Texture from Low-Resolution Images and Videos}}
\author{Xiangyu~Xu, Hao~Chen, Francesc~Moreno-Noguer, L{\'a}szl{\'o} A.~Jeni, Fernando~De la Torre%
\IEEEcompsocitemizethanks{
	\IEEEcompsocthanksitem
	X. Xu and L. Jeni are with Robotics Institute, Carnegie Mellon University, Pittsburgh, PA 15213, USA. E-mail: xuxiangyu2014@gmail.com, laszlojeni@cmu.edu.
	\IEEEcompsocthanksitem
	H. Chen is with Electrical and Computer Engineering, Carnegie Mellon University, Pittsburgh, PA 15213, USA. E-mail: hchen@cmu.edu.
	\IEEEcompsocthanksitem
	F. Moreno-Noguer is with Institut de Rob\`{o}tica i Inform\`{a}tica Industrial (CSIC-UPC), Barcelona, 08028, Spain. E-mail: fmoreno@iri.upc.edu.
	\IEEEcompsocthanksitem
	F. De la Torre is with Robotics Institute, Carnegie Mellon University, Pittsburgh, PA 15213, USA. He is also with Facebook Reality Lab (Oculus). E-mail: ftorre@cs.cmu.edu.
	}%
}

%

%
%
%
%
%
\IEEEtitleabstractindextext{%

\begin{abstract}
	3D human pose and shape estimation from monocular images has been an active research area in computer vision. Existing deep learning methods for this task rely on high-resolution input, which however, is not always available in many scenarios such as video surveillance and sports broadcasting. Two common approaches to deal with low-resolution images are applying super-resolution techniques to the input, which may result in unpleasant artifacts, or simply training one model for each resolution, which is impractical in many realistic applications.

	To address the above issues, this paper proposes a novel algorithm called RSC-Net, which consists of a Resolution-aware network, a Self-supervision loss, and a Contrastive learning scheme. The proposed method is able to learn 3D body pose and shape across different resolutions with one single model. The self-supervision loss enforces scale-consistency of the output, and the contrastive learning scheme enforces scale-consistency of the deep features. We show that both these new losses provide robustness when learning in a weakly-supervised manner. Moreover, we extend the RSC-Net to handle low-resolution videos and apply it to reconstruct textured 3D pedestrians from low-resolution input. Extensive experiments demonstrate that the RSC-Net can achieve consistently better results than the state-of-the-art methods for challenging low-resolution images.
\end{abstract}

}

\maketitle

\IEEEdisplaynontitleabstractindextext
\IEEEpeerreviewmaketitle

\begin{figure*}[t]
	\centering
	\includegraphics[width=0.85\textwidth]{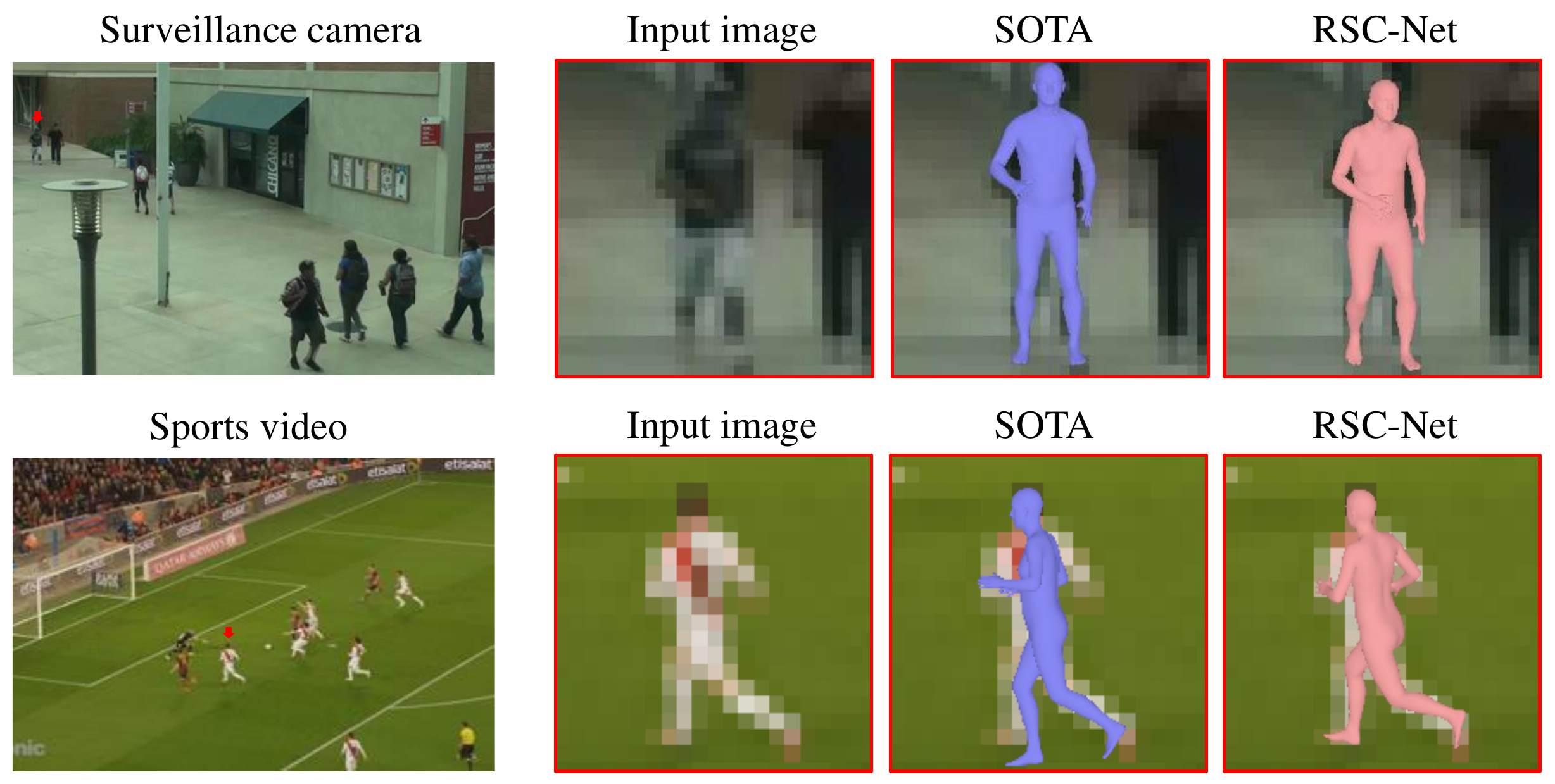}
	\vspace{-2mm}
	\caption{%
		3D human pose and shape estimation from real low-resolution images captured from a surveillance camera (top) and a sports video (bottom).
		SOTA method~\cite{kolotouros2019spin} that works well for high-resolution images performs poorly at low-resolution ones (note the orientation of the estimated human face and the posture of the arms and legs).
	}
	\label{fig:teaser_real}
\end{figure*}

\IEEEraisesectionheading{
\section{Introduction}\label{sec:intro}}
\IEEEPARstart{3}{D} human pose and shape estimation from 2D images is of great interest to the computer vision and graphics community.
Whereas significant progress has been made in this field,
it is often assumed that the input image is high-resolution and contains sufficient information for reconstructing the 3D human geometry in detail~\cite{alldieck2019learning,alldieck2018video,bogo2016keep,kanazawa2018end,kanazawa2019learning,kocabas2019vibe,kolotouros2019spin,natsume2019siclope,pavlakos2018learning,pumarola20193dpeople,saito2019pifu,zheng2019deephuman}.
However, this assumption does not always hold in practice, since people in applications of interest, such as surveillance cameras and sports videos, often have low resolutions~\cite{xxy-iccv17,wang2016studying,nishibori2014exemplar,neumann2018tiny,xu2019towards,oh2011large}.
As a result, existing algorithms trained with high-resolution images are prone to fail when applied to low-resolution inputs as shown in   Figure~\ref{fig:teaser_real}.

In this paper, we study the so-far unexplored problem  of estimating  3D human pose and shape from  low-resolution images where we find two major challenges.
First, the resolutions of the input images in real scenarios vary in a wide range, and a network trained for one specific resolution does not always work well for another.
One might consider overcoming this problem by simply training different models, one for each image resolution.
However, this is impractical in terms of memory and training computation. %
Alternatively, one could super-resolve the images to a sufficiently large resolution, but the super-resolution step often brings in unpleasant artifacts, which can result in significant domain gap between the super-resolved images and real high-resolution images, and thereby causes low reliability of 3D estimation.
To address this issue, we propose a resolution-aware deep neural network for 3D human pose and shape estimation that is robust to different image resolutions.
Our network builds upon two main components: a feature extractor shared across different resolutions and a set of resolution-dependent parameters to adaptively integrate the different-level features.

Another challenge we encounter is due to the fact that high-quality 3D annotations are hard to obtain, especially for in-the-wild data, and only a small portion of the training images have 3D ground truth labels~\cite{kanazawa2018end,kolotouros2019spin}, which brings difficulties for the training process.
Whereas most training images have 2D keypoint labels, they are usually not sufficient for training the network due to the inherent ambiguities in the 2D-to-3D mapping.
This weak-supervision problem is further accentuated in our task, as the low-resolution 3D estimation is not well constrained and has a large solution space due to limited pixel observations.
Therefore, directly training low-resolution models with weak supervision does not typically achieve good results.
Inspired by the self-supervised learning~\cite{laine2017temporal,tarvainen2017mean}, we propose a directional self-supervision loss to remedy the above issue.
Specifically, we enforce the consistency across the outputs of the same input image with different resolutions, such that the results of the higher-resolution images can act as guidance for lower-resolution input. We will show that this strategy significantly improves the 3D estimation results.

In addition to enforcing output consistency, we also devise an approach to enforce consistency of the feature representations across different resolutions.
Nevertheless, we find that the commonly-used mean squared error is not effective in measuring discrepancies between high-dimensional feature vectors.
Instead, we adapt the contrastive learning~\cite{oord2018representation,he2019momentum,chen2020simple} which aims to maximize the mutual information across the feature representations at different resolutions, and encourages the network to produce better features for the low-resolution input.

To summarize, we make the following contributions in this work.
First, we study the unexplored problem of 3D human pose and shape estimation from low-resolution images and present a simple yet effective solution called \netname{}, which is based on a novel resolution-aware network that can handle arbitrary-resolution input with one single model.
To the best of our knowledge, there is no prior work that solves this practically important problem.
Second, we propose a self-supervision loss to address the issue of weak supervision.
Third, we introduce contrastive learning for more effectively enforcing the consistency of the feature vectors across different resolutions.
Extensive experiments demonstrate that the proposed method outperforms the state-of-the-art algorithms on challenging low-resolution inputs and achieves robust performance for high-quality 3D human pose and shape estimation.

A preliminary version of this work is published in \cite{xu20203d}. Here we further extend the paper in the following aspects.
First, we provide a more thorough analysis of the proposed \netname{} with more qualitative examples and more quantitative evaluations on benchmark datasets.
Second, while the main contribution in this work is a single-frame model for 3D human pose and shape estimation, we also extend the \netname{} to handle low-resolution videos by adding a simple temporal post-processing step.
In addition, we show an important application of the \netname{}, \ie reconstructing textured 3D human from a low-resolution pedestrian image, by introducing an effective texture estimation network.

\section{Related Work}
We first review the state-of-the-art methods for 3D human pose and shape estimation and then discuss the low-resolution image recognition algorithms that are related to this work.

\subsection{3D Human pose and shape Estimation}
Recent years have witnessed significant progress in the field of 3D human pose and shape estimation from a single image~\cite{alldieck2019learning,alldieck2018video,alldieck2019tex2shape,bogo2016keep,doersch2019sim2real,kanazawa2018end,kanazawa2019learning,kocabas2019vibe,kolotouros2019spin,natsume2019siclope,pavlakos2018learning,pumarola20193dpeople,saito2019pifu,zheng2019deephuman,zhang2019predicting,zanfir2018monocular,moreno20173d,simo2013joint,simo20173d,omran2018neural}.
Existing methods for this task can be broadly categorized into two classes.
The first kind of approaches generally splits the 3D human estimation process into two stages: first transforming the input image into new representations, such as human 2D keypoints~\cite{bogo2016keep,pavlakos2018learning,natsume2019siclope,alldieck2018video,alldieck2019learning,doersch2019sim2real}, human silhouettes~\cite{pavlakos2018learning,alldieck2018video,natsume2019siclope}, body part segmentations~\cite{alldieck2019learning,omran2018neural}, UV mappings~\cite{alldieck2019tex2shape}, and optical flow~\cite{doersch2019sim2real}, and then regressing the 3D human parameters~\cite{loper2015smpl} from the transformed outputs of the last stage either with iterative optimization~\cite{bogo2016keep,alldieck2018video} or neural networks~\cite{pavlakos2018learning,alldieck2019learning,doersch2019sim2real,natsume2019siclope}.
\revise{For instance, Omran \etal propose a novel approach called Neural Body Fitting (NBF) which creatively integrates a statistical body model within a CNN, successfully leveraging both semantic body part segmentation and constraints of human bodies.}
As these methods map the original input images into simpler representation forms which are generally low-dimensional (\eg sparse~\cite{xu2020learning,Wright-Ma-2021}) and can be easily rendered, they can exploit a large amount of synthetic data (or data captured in constrained scenes) for training where there are sufficient high-quality 3D labels.
However, these two-stage systems are error-prone, as the errors from early stage may be accumulated or even deteriorated~\cite{kanazawa2018end}.
In addition, the intermediate results may throw away valuable information in the image such as context.
More importantly, the task of the first stage, \ie to estimate the intermediate representations, is usually difficult for low-resolution images, and thereby, the aforementioned two-stage models are not suitable to solve our problem of low-resolution 3D human pose and shape estimation.

Without relying on new representations, the second kind of approaches can directly regress the 3D parameters from the input image~\cite{kanazawa2018end,kanazawa2019learning,kolotouros2019spin,saito2019pifu,kocabas2019vibe,pumarola20193dpeople,zhang2019predicting}, where most methods are based on deep neural networks.
While being concise and not requiring the estimation of intermediate results, these methods usually suffer from the problem of weak supervision due to a lack of high-quality 3D ground truth.
Most existing works focus on this problem and have developed different techniques to solve it.
As a typical example, Kanazawa \etal \cite{kanazawa2018end} include a generative adversarial network (GAN)~\cite{gan} to constrain the solution space using the prior learned from 3D human data.
However, we find the GAN-based algorithm less effective for low-resolution input images where substantially fewer pixels are available.
Kolotouros~\cite{kolotouros2019spin} \etal integrate the optimization-based method \cite{bogo2016keep} into the training process of the deep network to more effectively exploit the 2D keypoints.
While achieving good improvements over~\cite{kanazawa2018end} on high-resolution images, \cite{kolotouros2019spin} cannot be easily applied to low-resolution input, as the low-resolution network cannot provide good initial results to start the optimization loop.
In addition, it significantly increases the training time.
Different from the above methods, we propose a low-resolution 3D human pose and shape estimation algorithm using a single image as input.
We propose a self-supervision loss and a contrastive feature loss which effectively remedy the problem of insufficient 3D supervision.

Note that temporal information has also been exploited to enforce the temporal consistency of the 3D estimation results~\cite{kanazawa2019learning,zhang2019predicting,kocabas2019vibe}, which however requires high-resolution video input.
In this work, we show that the proposed RSC-Net can be easily extended to exploit multi-frame information from low-resolution videos.

\subsection{Low-Resolution Image Recognition}
While there is no prior work for low-resolution 3D human pose and shape estimation, there are some related approaches to process low-resolution inputs for other image recognition tasks, such as 2D body pose estimation~\cite{neumann2018tiny}, face recognition~\cite{ge2018low,cheng2018low}, image classification~\cite{wang2016studying}, image retrieval~\cite{tan2018feature,Noh2019better}, and object detection~\cite{haris2018task,li2017perceptualgan}.
Most of these methods address the low-resolution issue by enhancing the degraded input, in either the image space~\cite{haris2018task,cheng2018low,wang2016studying} or the feature space~\cite{ge2018low,tan2018feature,li2017perceptualgan,Noh2019better}.
One typical image-space method~\cite{haris2018task} applies a super-resolution network which is trained to improve both the image quality (\ie per-pixel similarity such as PSNR) and the object detection performance.
However, the loss functions for higher PSNR and better recognition performance do not always agree with each other, which may lead to inferior solutions. Moreover, the super-resolution model may bring unpleasant artifacts (\eg over-smoothing and checkerboard patterns~\cite{xxy-iccv17}), resulting in domain gap between the super-resolved and the real high-resolution images.
Unlike the image enhancement based approaches, feature enhancement based methods~\cite{ge2018low,tan2018feature,li2017perceptualgan,Noh2019better} are not distracted by the image quality loss and thus can better focus on improving the recognition performance.
As a representative example, Ge \etal~\cite{ge2018low} use mean squared error (MSE) to enforce the similarity between the features of low-resolution and high-resolution images, obtaining good results for face recognition.
Different from the above approaches, Neumann~\etal~\cite{neumann2018tiny} propose a novel method for low-resolution 2D body pose estimation by predicting a probability map with Gaussian Mixture Model,
which, however, cannot be easily extended to 3D human pose and shape estimation.
In this work, we apply the feature enhancement strategy to low-resolution 3D human pose and shape estimation.
Instead of using MSE for measuring feature similarity, we introduce the contrastive learning~\cite{oord2018representation} which can more effectively maximize the mutual information across the features of different resolutions.
In addition, we handle different-resolution inputs with a resolution-aware neural network.

\begin{figure*}[t]
	\footnotesize
	\begin{center}
		\begin{tabular}{c}
			\includegraphics[width = 0.99\linewidth]{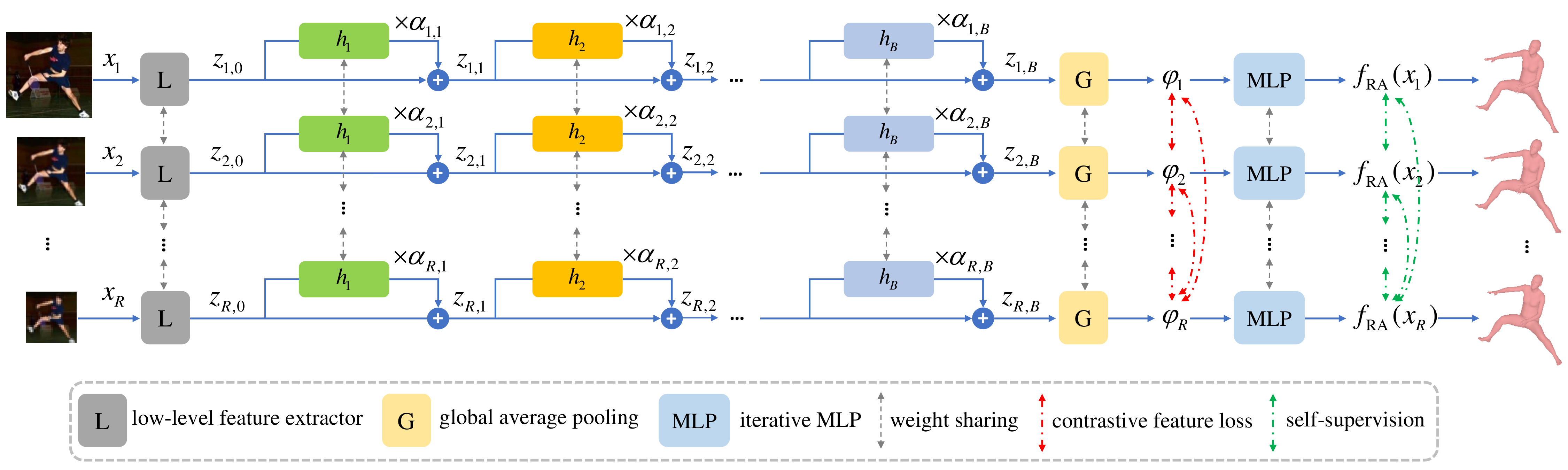} \\
		\end{tabular}
	\end{center}
	\vspace{-2mm}
	\caption{Overview of the proposed \netname{}.
		The resolution-aware network $f_{\text{RA}}$ is trained with a combination of the basic loss (omitted in the figure for simplicity), self-supervision loss and contrastive feature loss. The modules with the same colors are shared across different resolutions, while the matrix $\alpha$ is resolution-dependent.
		$\{x_i\}_{i=1}^R$ represents the different resolution versions of the same image, where $R$ is the number of resolutions considered in this work.
		We resize the different resolution inputs with bicubic interpolation before feeding them into the network.
		Note that in real implementation, we split the $R$ resolutions into $P$ ranges to ease the training burden (see the text in Section~\ref{sec:res-aware}).
		During the training phase, we jointly train the \netname{} with all different resolution images.
		During inference, we first decide the resolution range of the input image and then choose the suitable row of parameters in $\alpha$ for usage in the network.
	}
	\label{fig:resolution-aware network}
\end{figure*}
\section{Algorithm}
We study the problem of 3D human pose and shape estimation from a single low-resolution image.
We propose a simple yet effective solution (RSC-Net) for this problem, which consists of a resolution-aware network and new training losses. An overview of the RSC-Net is shown in Figure~\ref{fig:resolution-aware network}.
In this section, we first introduce the 3D human representation model (Section~\ref{sec:smpl}) and the baseline network for 3D human estimation from a single 2D image (Section~\ref{sec:res-aware}).
We then describe the proposed resolution-aware model (Section~\ref{sec:res-aware}) as well as the self-supervision loss (Section~\ref{sec:SS}) and the contrastive learning strategy for training the network (Section~\ref{sec:contra}).
In addition, we introduce a temporal recurrent module to extend the proposed single-frame model to low-resolution videos (Section~\ref{sec:video}).
We also show how the RSC-Net can be used to reconstruct textured human from a low-resolution pedestrian image (Section~\ref{sec:texture}).

\subsection{3D Human pose and shape Representation} \label{sec:smpl}
We represent the 3D human body using the Skinned Multi-Person Linear (SMPL) model~\cite{loper2015smpl}.
The SMPL is a parametric model which describes the body pose and shape with two sets of parameters $\beta$ and $\theta$, respectively.
The body shape is represented by a basis in a low-dimensional shape space learned from a training set of 3D human scans, and the parameters $\beta \in \mathbb{R}^{10}$ are coefficients of the basis vectors.
The body pose is defined by a skeleton rig with $K=24$ joints including the body root, and the pose parameters $\theta\in \mathbb{R}^{3K}$ are the axis-angle representations of the relative rotation between different body parts as well as the global rotation of the body root.
With $\beta$ and $\theta$, we can obtain the 3D body mesh:
\begin{align}
	M = f_{\text{SMPL}}(\beta, \theta),
\end{align}
where $M \in \mathbb{R}^{N \times 3}$ is a triangulated mesh with $N=6890$ vertices.

Similar to prior works~\cite{kanazawa2018end,kolotouros2019spin}, we can predict the 3D locations of the body joints $X$ with the body mesh using a pre-trained mapping matrix $W \in \mathbb{R}^{K \times N}$:
\begin{align} \label{eq:3D_joints}
	X \in \mathbb{R}^{K \times 3} = WM.
\end{align}
With the 3D human joints, we can then use a perspective camera model to project the body joints from 3D to 2D. Assuming the camera parameters are $\delta \in \mathbb{R}^{3}$ which define the 3D translation of the camera, the 2D keypoints can be written as:
\begin{align} \label{eq:2D_joints}
	J \in \mathbb{R}^{K \times 2} = f_{\text{project}}(X, \delta),
\end{align}
where $f_{\text{project}}$ is the perspective projection function~\cite{hartley2003multiple}. %

\subsection{Resolution-Aware 3D Human Estimation} \label{sec:res-aware}

\vspace{1mm}
\noindent \textbf{Baseline network.}
Similar to the existing methods~\cite{kanazawa2018end,kolotouros2019spin}, we use the deep convolutional neural network (CNN) for 3D human estimation, where the ResNet-50~\cite{resnet} is employed to extract features from the input image.
The building block of the ResNet (\ie ResBlock \cite{he2016identity}) can be formulated as:
\begin{align} \label{eq:resblock}
	z_{k} &= z_{k-1} + h_{k}(z_{k-1}),
\end{align}
where $z_{k}$ represents the output feature maps of the $k$-th ResBlock, and $h_k$ is the nonlinear function used to learn the feature residuals, which is modeled by several convolutional layers with ReLU activation~\cite{nair2010rectified}.
The ResNet stacks $B$ ResBlocks together, and recursively we will have: 
\revise{\begin{align}\label{eq:resnet}
	z_{B} &= z_{B-1} + h_{B}(z_{B-1}) \nonumber \\
	&= z_{B-2} + h_{B-1}(z_{B-2}) + h_{B}(z_{B-1}) \nonumber \\
	&= z_{B-3} + h_{B-2}(z_{B-3}) + h_{B-1}(z_{B-2}) + h_{B}(z_{B-1}) \nonumber \\
	&... \nonumber \\
	&= z_{0} + \sum_{k=1}^B h_{k}(z_{k-1}), 
\end{align}}\\
where $z_0$ is the low-level feature maps extracted from the input image $x$ with convolutional layers, and $z_B$ is a combination of different level residual maps from all the ResBlocks.
Note that we do not explicitly consider the downsampling ResBlocks in Eq.~\ref{eq:resnet} for clarity.
With the output features of the ResNet, we can use global average pooling to obtain a feature vector $\varphi$ and employ an iterative MLP for regressing the 3D parameters $\beta, \theta, \delta$ similar to \cite{kanazawa2018end,kolotouros2019spin}. \revise{Specifically, the iterative MLP regresses the parameters in an iterative error feedback loop, where progressive changes are made to the current estimate. Thus, the network can learn residuals to recurrently refine the initial output, leading to more accurate results. We refer to \cite{kanazawa2018end} for more details.}

\vspace{1mm}
\noindent \textbf{Resolution-aware network.}
The baseline network is originally designed for high-resolution images with input size of 224$\times$224 pixels,
whereas the image resolutions for human in real scenarios can be much lower and vary in a wide range.
A straightforward way to deal with these low-resolution inputs is to train different networks for all possible resolutions and choose the suitable one for each test image.
However, this is not practical for real applications given the
high computational complexity and memory requirement.

To solve this problem, we propose a resolution-aware network, and the main idea is that the different-resolution images with the same contents are largely similar as shown in Figure~\ref{fig:resolution-aware network}, and can share most parts of the feature extractor. And only a small amount of parameters are needed to be resolution-dependent to account for the characteristics of different image resolutions.
Towards this end, instead of directly combining the different level features as in \textcolor{blue}{Eq.}~\ref{eq:resnet},
we learn a matrix $\alpha$ to adaptively fuse the residual maps from the ResBlocks for each resolution as shown in Figure~\ref{fig:resolution-aware network}. This allows different resolution inputs to have tailored feature extractors for better 3D estimation.
Specifically, we formulate the output of the proposed resolution-aware network as:
\begin{align} \label{eq:res_aware_net}
	z_{i,B} &= z_{i,0} + \sum_{k=1}^B \alpha_{i,k} h_{k}(z_{i,k-1}),~~~i=1,2,\dots,R,
\end{align}
where $i$ is the index for different image resolutions, and larger $i$ indicates smaller image. $i=1$ corresponds to the original high-resolution input. $\alpha \in \mathbb{R}^{R \times B}$, where $R$ denotes the number of all the image resolutions considered in this work.
$z_{i,k}$ and $\alpha_{i,k}$ respectively represent the output and the fusion weight of the $k$-th ResBlock for the $i$-th input resolution.
According to Eq.~\ref{eq:res_aware_net}, the original ResBlock in Eq.~\ref{eq:resblock} is modified as: %
\begin{align} \label{eq:res_aware_block}
z_{i,k} &= z_{i,k-1} + \alpha_{i,k} h_{k}(z_{i,k-1}).
\end{align}
Note that we use a slightly different notation here compared with Eq.~\ref{eq:resblock} and Eq.~\ref{eq:resnet} which do not have the index $i$ for image resolution, as the baseline network is not resolution-aware and applies the same operations to different resolution inputs.
This new formulation of the ResBlock in Eq.~\ref{eq:res_aware_block} is in form similar to the concurrent work~\cite{bachlechner2020rezero}, and the difference is that we use $\alpha$ for resolution-aware feature extractors while \cite{bachlechner2020rezero} uses it for fast convergence in training networks.

To train the above network, we need to conduct the downsampling operation to each high-resolution image in the training dataset for $R-1$ times, such that each row of parameters in $\alpha$ have their corresponding training data.
Whereas the original training datasets~\cite{3dpw,human3.6,mpi-inf-3dhp,mpii,coco} are already quite large for the diversity of the training images, it will be further augmented by $R-1$ times, which significantly increases the computational burden of the training process.
To alleviate the training issues, we divide all the $R$ resolutions %
into $P$ ranges and only learn one set of parameters for each range.
We design the first resolution range to only have the original high-resolution image, and for the other ranges, we randomly sample a resolution in each range during each training iteration.
The training images with different resolutions can be denoted as $\{x_i\}_{i=1}^P$  where the smaller images $x_2, x_3, \dots, x_P$ are synthesized from the same high-resolution image $x_1$ with bicubic interpolation.
With this strategy, the training set can be much smaller without losing diversity, and we can have a lower-dimensional matrix $\alpha \in \mathbb{R}^{P \times B}$, where the number of parameters can be reduced from $RB$ to $PB$.
During inference, we first decide the resolution range of the input image and then choose the suitable row of parameters in $\alpha$ for usage in the network.

\vspace{1mm}
\noindent \textbf{Progressive training.}
Directly using different resolution images for training all at once can lead to difficulties in optimizing the proposed model since the network needs to handle inputs with complex resolution properties simultaneously.
Instead, we train the proposed network in a progressive manner, where the higher-resolution images are easier to handle and thus first processed in training, and more challenging ones with lower resolutions are subsequently added.
In this way, we alleviate the difficulty of the training process and the proposed model can evolve progressively.
Note that this strategy is related to the curriculum learning algorithm~\cite{bengio2009curriculum} in that difficult examples are gradually emphasized.
However, \cite{bengio2009curriculum} does not explicitly consider multi-resolution training.

\vspace{1mm}
\noindent \textbf{Basic loss function.}
Similar to the previous works~\cite{kanazawa2018end,kolotouros2019spin}, the basic loss of our network is a combination of 3D and 2D losses.
Suppose the output of the proposed network for input image $x_i$ is $ [ \hat{\beta}_i, \hat{\theta}_i, \hat{\delta}_i ] = f_{\text{RA}}(x_i)$ where $i$ is the resolution index, and ${X_g, J_g, \beta_{\text{g}}}, {\theta_{\text{g}}}$ are the ground truth 3D and 2D keypoints and SMPL parameters.
The basic loss function can be written as:
\begin{align}\label{eq:basic loss}
	L_{\text{b}} =&\sum_i \| [\hat{\beta}_i, \hat{\theta}_i] - [{\beta_{\text{g}}}, {\theta_{\text{g}}}] \|_2^2 \nonumber \\  &+ \lambda_1 \| \hat{X}_i - X_{\text{g}} \|_2^2  +\lambda_2 \| \hat{J}_i - J_{\text{g}} \|_2^2,
\end{align}
where $\hat{X}_i$ and $\hat{J}_i$ are estimated with Eq.~\ref{eq:3D_joints} and Eq.~\ref{eq:2D_joints}, respectively. $\lambda_1$ and $\lambda_2$ are hyper-parameters for balancing different terms.
Note that while all the training images have 2D keypoint labels $J_g$ in  Eq.~\ref{eq:basic loss}, only a limited portion of them have 3D ground truth $X_g, \beta_g, \theta_g$.
For the training images without 3D labels, we simply omit the first two terms in Eq.~\ref{eq:basic loss} similar to \cite{kanazawa2018end,kanazawa2019learning,kolotouros2019spin}

\subsection{Self-Supervised Loss} \label{sec:SS}
The 3D human pose and shape estimation is usually posed as a weakly-supervised problem as only a small part of the training data has 3D labels, and this is especially the case for in-the-wild images where accurate 3D annotations cannot be easily captured.
This issue gets even worse for the low-resolution images, as the 3D estimation is not well constrained by limited pixel observations and requires strong supervision signal during training to find a good solution.

To remedy this problem, we propose a self-supervision loss to assist the basic loss for training the resolution-aware network $f_\text{RA}$.
This new loss term is inspired by the self-supervised learning algorithm~\cite{laine2017temporal} which improves the training by minimizing the MSE between the network predictions under different input augmentation conditions.
For our problem, we naturally have the same input with different data augmentations, \ie the different-resolution images synthesized from the same high-resolution image.
Thus, the self-supervision loss can be formulated by enforcing the consistency across the outputs of different image resolutions:
\begin{align} \label{eq:self-supervision-v1}
	\sum_{i, j}	\|f_{\text{RA}}(x_i)-f_{\text{RA}}(x_j)\|_2^2.
\end{align}

However, a major difference between our work and the original self-supervision method~\cite{laine2017temporal} is that we are generally more confident in the predictions of the higher-resolution images, while \cite{laine2017temporal} treats the results under different input augmentations equally.
To exploit this prior knowledge, we improve the loss in Eq.~\ref{eq:self-supervision-v1} and propose a directional self-supervision loss:
\begin{align}
	\begin{aligned} \label{eq:self-supervision-v2}
		L_{\text{s}} =& \sum_{i,j} w_{i,j}	\|\bar{f}_{\text{RA}}(x_i)-f_{\text{RA}}(x_j)\|_2^2, \\
		w_{i,j} &= \mathbbm{1}(j-i>0) \cdot (j-i),
	\end{aligned}
\end{align}
where $w_{i,j}$ is the loss weight for an image pair $(x_i, x_j)$, and it is nonzero only when $x_i$ has higher-resolution than $x_j$.
$\bar{f}_{\text{RA}}$ represents a fixed network, and the gradients are not back-propagated through it such that the lower-resolution image $x_j$ is encouraged to have similar output to higher-resolution $x_i$ but not vice versa.
In addition, since higher-resolution results usually provide higher-quality guidance during training, we give a larger weight to larger resolution difference by the term $(j-i)$ in $w_{i,j}$.
Note that we use all the resolutions that are higher than $x_j$ as supervision in Eq.~\ref{eq:self-supervision-v2} instead of only using the highest resolution $x_1$, as the results of $x_j$ and $x_1$ can differ from each other significantly for a large $j$, and the results of the resolutions between $x_j$ and $x_1$ can act as soft targets during training. In \cite{hinton2015distilling}, Hinton \etal show the effectiveness of the ``dark knowledge'' in soft targets, and similarly for low-resolution 3D human pose and shape estimation, we also find that it is important to provide the challenging input a hierarchical supervision signal such that the learning targets are not too difficult for the network to follow.

\subsection{Contrastive Learning} \label{sec:contra}
While the self-supervision loss enforces the consistency of the network outputs across different image resolutions, we can further improve the model training by encouraging the consistency of the final feature representation $\varphi$ encoded by the network, such that features of lower-resolution images are closer to those of higher-resolution images.
Similar to Eq.~\ref{eq:self-supervision-v2}, we have the feature consistency loss:
\begin{align} \label{eq:feature loss}
	L_{\text{f}} = \sum_{i,j} w_{i,j} g(\bar{\varphi}_i, \varphi_j),
\end{align}
where $\varphi_{i}$ is the feature vector of the $i$-th resolution input image $x_i$, and $\bar{\varphi}$ denotes a fixed feature vector without gradient back-propagation.
$w_{i,j}$ is identical to that in Eq.~\ref{eq:self-supervision-v2}.
The function $g$ is used to measure the distance between two feature vectors, and a straightforward choice is the MSE as in Eq.~\ref{eq:self-supervision-v2}.
However, the extracted features $\varphi$ usually have very high dimensions (\eg 2048), and the MSE loss is not effective in modeling correlations of the complex structures in high-dimensional representations, due to the fact that it can be decomposed element-wisely, \ie assuming independence between elements in the feature vectors~\cite{oord2018representation,tian2019contrastive}.
Moreover, unimodal losses such as MSE can be easily affected by noise or insignificant structures in the features, and it would be desirable to have a loss function exploiting more global structures~\cite{oord2018representation}.

Towards this end, we propose a contrastive feature loss similar to \cite{oord2018representation,chen2020simple,he2019momentum,tian2019contrastive} to maximize the mutual information across the feature representations of different resolutions.
The main idea behind our contrastive loss is to encourage the feature representation to be close for the same sample with different resolutions and far for different image samples.
Mathematically, the contrastive function can be written as:
\begin{align}\label{eq:contrastive}
	g(\bar{\varphi_{i}}, \varphi_{j}) = -\log \frac{\exp(s(\bar{\varphi_{i}}, \varphi_{j}) / \tau)}{\exp(s(\bar{\varphi_{i}}, \varphi_{j}) / \tau) + \sum_{q \in \mathcal{Q}} \exp(s(q, \varphi_{j}) / \tau) },
\end{align}
where $s$ represents the cosine similarity function, and $\tau$ is a temperature hyper-parameter~\cite{chen2020simple}.
$\varphi_{i}, \varphi_{j}$ are the features of the same input with different resolutions.
$\mathcal{Q}$ is a queue of data samples, which is constructed and progressively updated during training, and $\varphi_i,\varphi_j \notin \mathcal{Q}$.
We use a method similar to \cite{he2019momentum} to update the queue, \ie after each iteration, the current mini-batch is enqueued, and the oldest mini-batch in the queue is removed.
Supposing the size of the queue is $|\mathcal{Q}|$,
the contrastive loss is essentially a $(|\mathcal{Q}|+1)$-way softmax-based classifier which classifies different resolutions $(\varphi_i, \varphi_j)$ as a positive pair while different contents $(q, \varphi_j)$ as a negative pair.
As the feature extractor of the higher resolution image does not have gradients in Eq.~\ref{eq:contrastive}, the proposed loss function enforces the network to generate higher-quality features for the low-resolution input image.

Our final loss is a combination of the basic loss, self-supervision loss, and  contrastive feature loss: $L_{\text{b}} + \lambda_\text{s} L_\text{s} + \lambda_\text{f} L_\text{f}$, where $\lambda_\text{s}$ and $\lambda_\text{f}$ are hyper-parameters.
\begin{figure}[t]
	\footnotesize
	\begin{center}
		\begin{tabular}{c}
			\includegraphics[width = 0.95\linewidth]{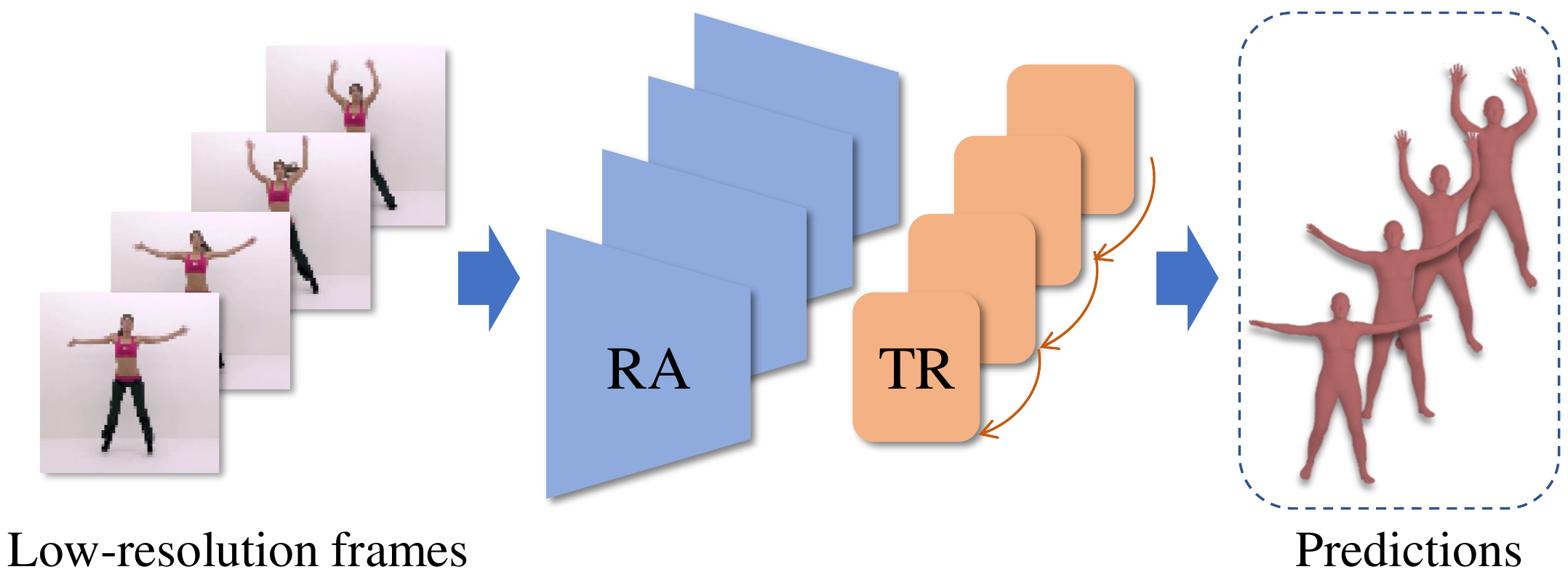} \\
		\end{tabular}
	\end{center}
	\vspace{-2mm}
	\caption{Extension for handling video input. We first use the proposed RSC-Net to extract features for each frame and then apply a Temporal Recurrent module (TR)~\cite{kocabas2019vibe} to improve the per-frame results.
	}
	\label{fig:vibe_network}
\end{figure}

\subsection{Extension for Handling Video Input} \label{sec:video}
While this work mainly focuses on 3D human pose and shape estimation from a single low-resolution image, it is a common practice to exploit the temporal information to further improve the predictions when videos are available~\cite{kanazawa2019learning,kocabas2019vibe}.
In this section, we show that we can easily extend the RSC-Net to  handle low-resolution video input by incorporating a temporal post-processing step.

Specifically, we employ a strategy similar to the VIBE algorithm~\cite{kocabas2019vibe}, which first performs per-frame estimation of the 3D human pose and shape with a pre-trained single-image model and then uses a temporal recurrent module composed of a Gated Recurrent Unit (GRU)~\cite{choetal2014gru} and an MLP to aggregate multi-frame information.
We adapt this method to deal with low-resolution videos by adding the temporal recurrent module to the proposed resolution-aware network, as shown in Figure~\ref{fig:vibe_network}.
We first use the pre-trained resolution-aware network to extract feature vectors $\varphi$ from each frame of the input video, and then apply the recurrent module to the features to encode the temporal relations across frames, which facilitates generating more accurate and temporally-coherent SMPL outputs.
Similar to \cite{kocabas2019vibe}, we also use a GAN loss to train the temporal recurrent module, where a motion discriminator takes a sequence of SMPL predictions as input and learns to discriminate between generated motion and ground-truth motion.
Such an adversarial learning scheme \cite{gan} enforces more constraints on the ill-posed low-resolution estimation problem and enhances both the consistency and feasibility of the generated motion sequences.

\begin{figure*}[t]
	\footnotesize
	\begin{center}
		\begin{tabular}{c}
			\includegraphics[width = 0.95\linewidth]{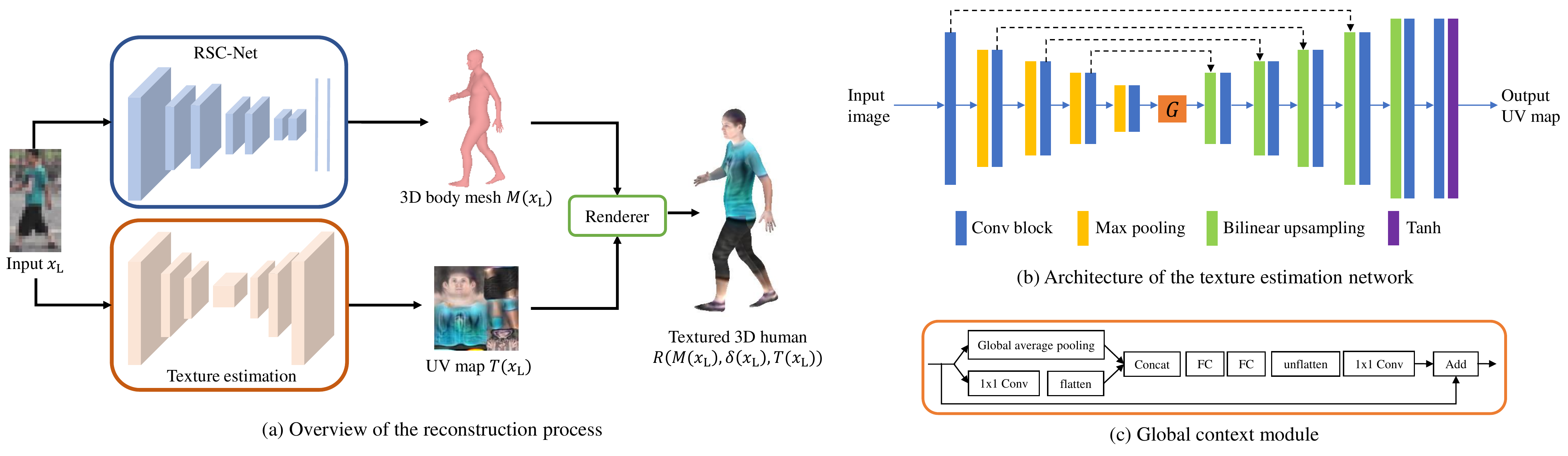} \\
		\end{tabular}
	\end{center}
	\vspace{-2mm}
	\caption{Textured 3D human reconstruction from a single low-resolution image. We show an overview of the reconstruction process in (a). Note that the \netname{} in (a) also outputs the camera parameter $\delta(x_\text{L})$ which is omitted in the figure for clarity.
	The detailed architecture of the proposed texture estimation network is illustrated in (b).
	The main difference between the baseline texture estimation network~\cite{wang2019re} and our model is the global context module $G$ detailed in (c). The ``flatten'' in (c) represents the feature flattening layer which reshapes the feature map into a one-dimensional vector. ``unflatten'' is the inverse-operation reshaping the global feature vector into feature maps. ``FC'' is the fully-connected layer.
	}
	\label{fig:tex_overview}
		\vspace{-3mm}
\end{figure*}
\begin{figure}[t]
	\footnotesize
	\begin{center}
		\begin{tabular}{c}
			\includegraphics[width = 0.95\linewidth]{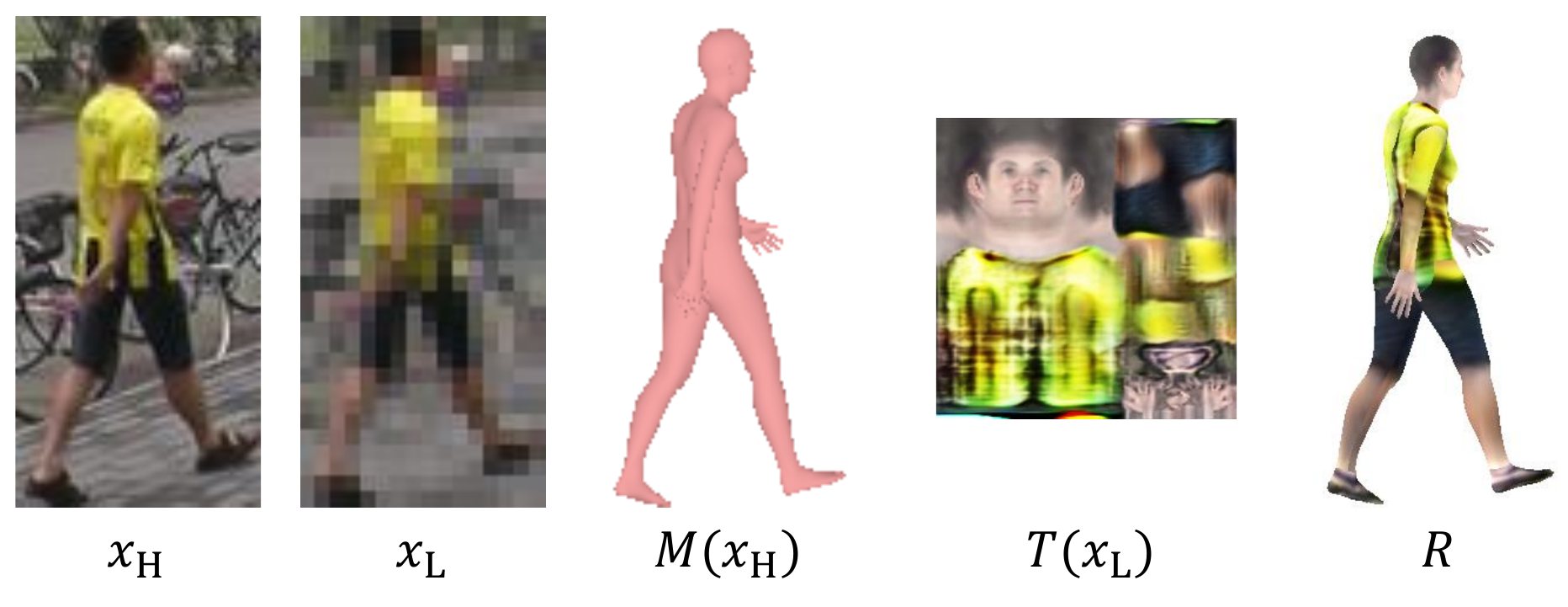} \\
		\end{tabular}
	\end{center}
	\vspace{-2mm}
	\caption{Intermediate results for training the texture estimation network. $x_\text{H}$ is a high-resolution image from the training dataset, and $x_\text{L}$ is the corresponding low-resolution version synthesized with bicubic interpolation. $M$ and $T$ represent the predicted 3D body mesh and UV map. 
	$R$ is the textured output rendered with Pytorch3D~\cite{ravi2020pytorch3d}.
	}
	\label{fig:opendr}
		\vspace{-3mm}
\end{figure}

\begin{figure*}[t]
	\centering
	\includegraphics[width=0.93\textwidth]{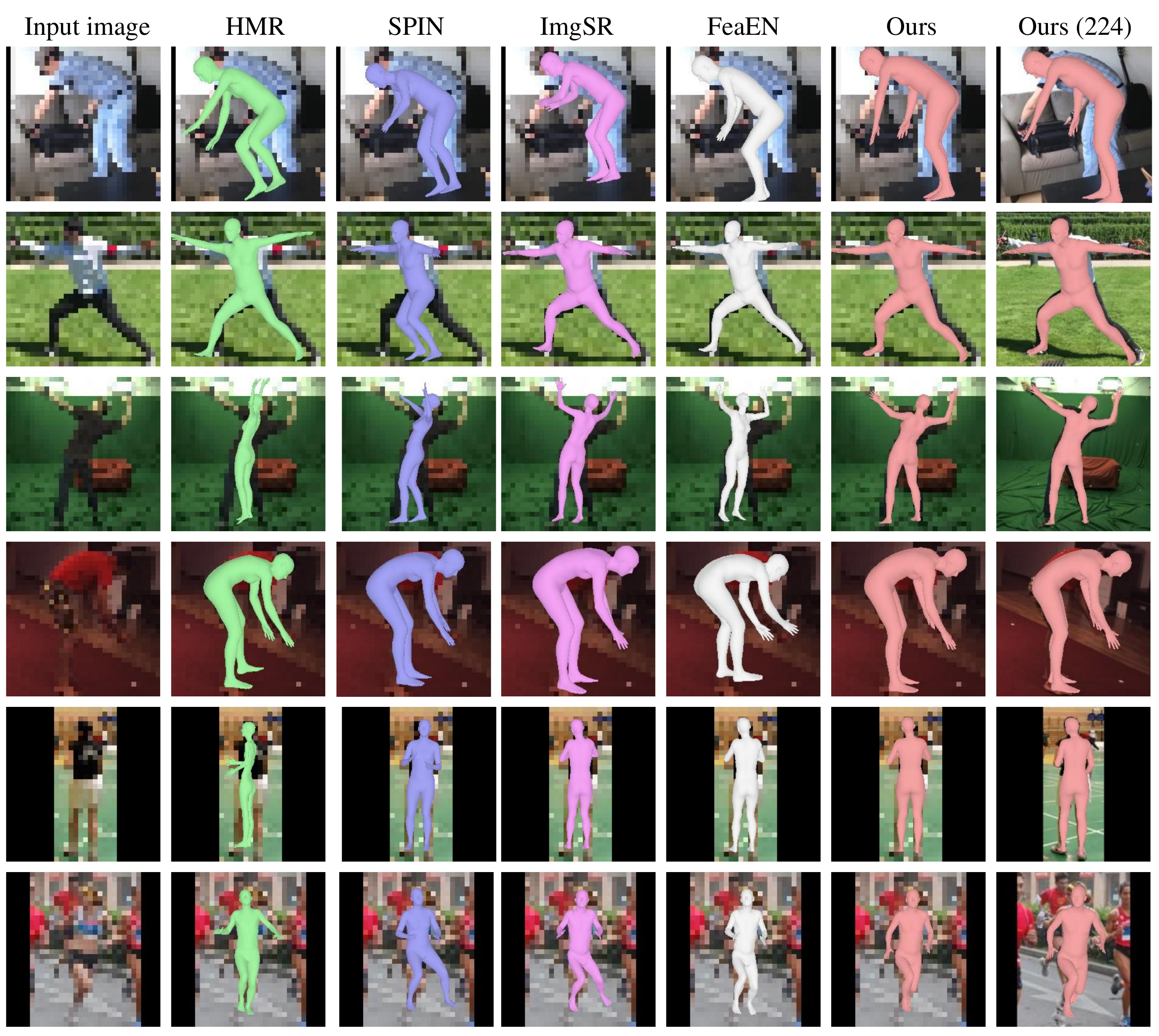}
	\caption{Visual comparisons with the state-of-the-art methods on challenging low-resolution input. The input image has a resolution of $32\times32$. The results of high-resolution images are also included as references.
	}
	\label{fig:SOTA comparison}
\end{figure*}

\subsection{Texture Reconstruction for Low-Resolution Human} \label{sec:texture}
To further demonstrate the potential of our approach, we propose a human texture estimation network which, together with the RSC-Net, can be used to reconstruct textured 3D human body from a single low-resolution pedestrian image (an overview is shown in Figure~\ref{fig:tex_overview}(a)).

\vspace{1mm}
{\bf \noindent Network structure.}
Similar to the state-of-the-art human texture estimation method~\cite{wang2019re} which is originally designed for high-resolution input, we use an encoder-decoder framework with concatenative skip-connections to predict the desired UV map from low-resolution images as shown in Figure~\ref{fig:tex_overview}(b),
where we add an extra upsampling layer to the baseline model of \cite{wang2019re} to encourage more details in the reconstructed UV map.

While the basic encoder-decoder model of \cite{wang2019re} works reasonably well in practice, it has the potential problem that the fully convolutional structure only exploits the information from a local receptive field.
Note that the contents of the input image of the texture network and the output UV map are significantly misaligned, \eg the legs in the bottom of the input are located at the top of the UV map as shown in Figure~\ref{fig:tex_overview}(a).
A successful solution to this problem should be able to grasp the global context information such that it can better spatially re-arrange the output pixels.
To this end, we introduce a global context module in Figure~\ref{fig:tex_overview}(c), where we combine two types of mechanisms for global information aggregation, \ie global averaging pooling~\cite{resnet}, and feature flattening~\cite{depth_eigen2014} which reshapes the input feature maps into a one-dimensional vector.
The feature vectors from the pooling and flattening layers are first concatenated and then processed by two fully-connected layers such that the information from different spatial locations can be aggregated and better exploited for UV map prediction.
As the input feature maps of the global context module have quite high dimensionality, we use a 1$\times$1 convolution layer to reduce the feature maps before sending them to the flattening layer.
The feature dimensionality is recovered with another 1$\times$1 convolution layer after the unflattening layer in Figure~\ref{fig:tex_overview}(c) which reshapes the processed vectors back to feature maps.
We finally fuse the unflattened global image features back to the input features with point-wise summation.
Note that we also tried to fuse the features with a concatenation layer which, however, does not work well in our experiments.

\vspace{1mm}
{\bf \noindent Training process.}
With a high-resolution human image $x_\text{H}$ (Figure~\ref{fig:opendr}) from the training dataset, we can synthesize a low-resolution image $x_\text{L}$ with bicubic interpolation.
We first use the high-resolution image $x_\text{H}$ as input of the proposed \netname{} to obtain a high-quality 3D body mesh $M(x_\text{H})$ as well as its corresponding camera parameters $\delta(x_\text{H})$.
Then we predict the UV map $T(x_\text{L})$ for the low-resolution image with the texture estimation network,
and the textured human image can be reconstructed by $R(M(x_\text{H}), \delta(x_\text{H}), T(x_\text{L}))$ using a differentiable renderer $R$.
Different from \cite{wang2019re} where the OpenDR~\cite{loper2014opendr} is used as the renderer, we employ the more modern tool Pytorch3D~\cite{ravi2020pytorch3d} for differentiable rendering.
Finally, the texture estimation network can be trained with the Re-identification (ReID) loss of \cite{wang2019re}:
\begin{align} \label{eq:reid}
\| f_\text{Re}( R(M(x_\text{H}), \delta(x_\text{H}), T(x_\text{L})) ) - f_\text{Re}( x_\text{H} ) \|_2,
\end{align}
where $f_\text{Re}$ represents a pre-trained ReID network that is explicitly trained to minimize the distance between images of the same identity.
As the human identity is mostly characterized by the body textures, minimizing the ReID loss can effectively encourage the rendered output to have similar appearance as the high-resolution image $x_\text{H}$, while alleviating the effect of the misalignment between the ground truth and output. During inference for a low-resolution image, the textured human can be reconstructed by $R(M(x_\text{L}), \delta(x_\text{L}), T(x_\text{L}))$ as show in Figure~\ref{fig:tex_overview}(a).

\section{Experiments}
We first describe the implementation details of the proposed algorithm.
Then we compare the \netname{} with the state-of-the-art 3D human pose and shape estimation approaches on different benchmark datasets.
In addition, we also provide evaluations on low-resolution video input.
We finally present quantitative and qualitative results to evaluate the proposed texture estimation network.

\begin{figure}[t]
	\centering
	\includegraphics[width=0.93\linewidth]{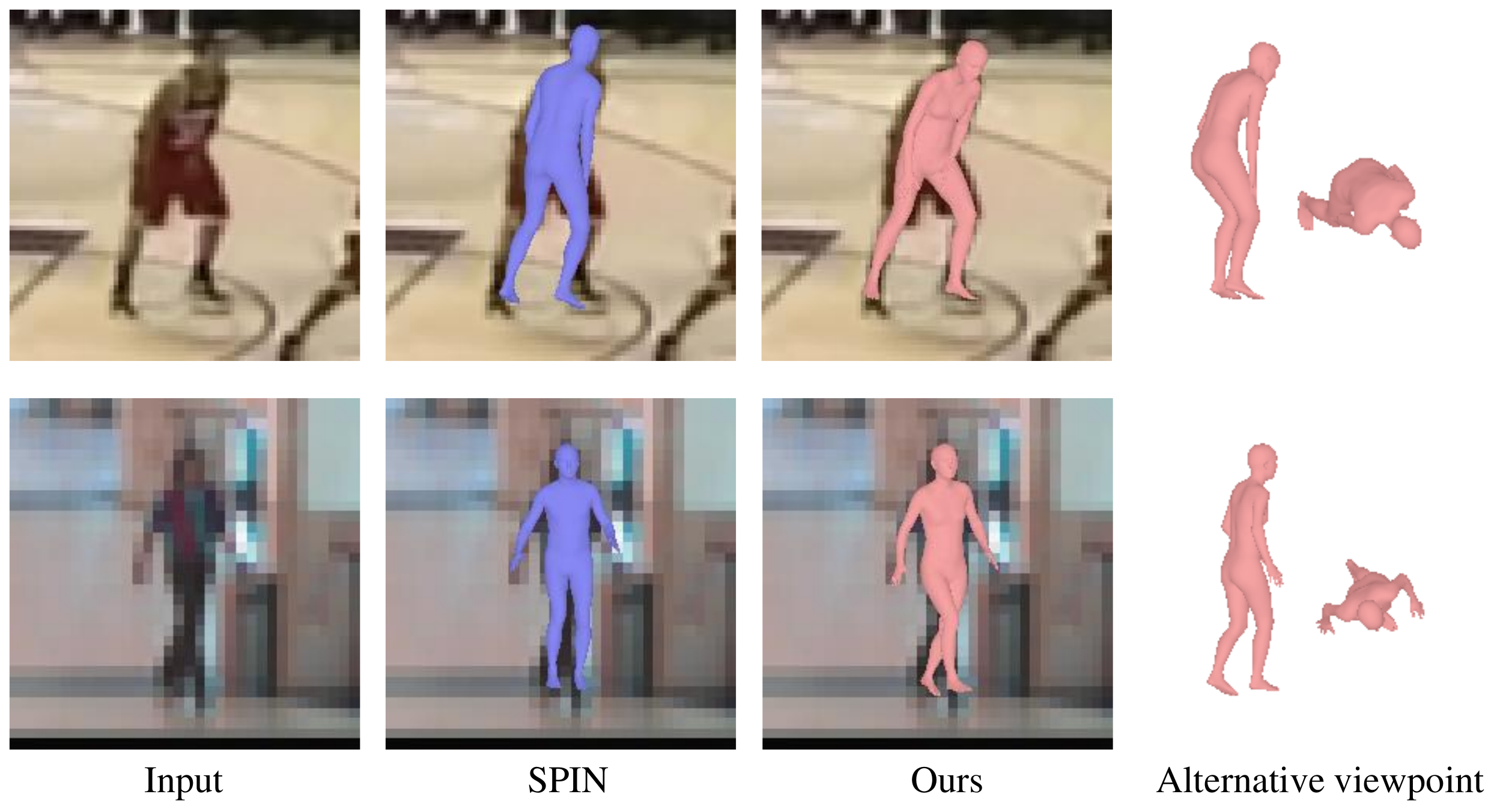}
	\vspace{-2mm}
	\caption{\revise{Visual examples of real low-resolution images captured from the Internet.
		SOTA method~\cite{kolotouros2019spin} that works well for high-resolution images performs poorly at low-resolution ones.} }
	\vspace{-3.5mm}
	\label{fig:real}
\end{figure} 

\begin{table*}[t]
	\centering
	\vspace{0mm}
	\caption{\label{tab:3dpw-mpi-h36m}
		Results on the 3DPW~\cite{3dpw}, MPI-INF-3DHP~\cite{mpi-inf-3dhp}, and H36M~\cite{human3.6} datasets. The proposed method compares favorably against the baseline approaches for challenging low-resolution (32$\times$32) images.}
	\vspace{-2mm}
	\begin{tabular}{l*{2}{>{\centering\arraybackslash}p{0.09\textwidth}} >{\centering\arraybackslash}p{0.002\textwidth} *{2}{>{\centering\arraybackslash}p{0.09\textwidth}} >{\centering\arraybackslash}p{0.002\textwidth} *{2}{>{\centering\arraybackslash}p{0.09\textwidth}} }
		\toprule
		\multirow{3}{*}{Methods~~} & \multicolumn{2}{c}{3DPW}           &               & \multicolumn{2}{c}{MPI-INF-3DHP}    &               & \multicolumn{2}{c}{H36M}                       \\
		\cmidrule{2-3} \cmidrule{5-6} \cmidrule{8-9}
		& \multicolumn{1}{c}{MPJPE} & \multicolumn{1}{c}{MPJPE-PA} & & \multicolumn{1}{c}{MPJPE} & \multicolumn{1}{c}{MPJPE-PA} & & \multicolumn{1}{c}{MPJPE} & \multicolumn{1}{c}{MPJPE-PA} \\
		\midrule
		HMR                      & 142.29          & 79.73    & &  133.25          & 85.30  & & 93.08          & 63.96        \\
		SPIN                     & 137.61         & 78.44    & &  127.27         & 83.38  & & 86.17          & 57.35         \\
		ImgSR                    & 146.58         & 81.07    & &   125.91         & 83.52  & & 87.23          & 57.06       \\
		FeaEN                    & 143.51          & 77.21   & &   124.99          & 81.80  & & 93.88          & 62.70      \\
		Ours                     & \textbf{117.12} & \textbf{67.59} && \textbf{115.80} & \textbf{78.68}  & & \textbf{79.59} & \textbf{53.56} \\
		\bottomrule
	\end{tabular}
\end{table*}

\begin{figure*}[t]
	\centering
	\includegraphics[width=0.93\textwidth]{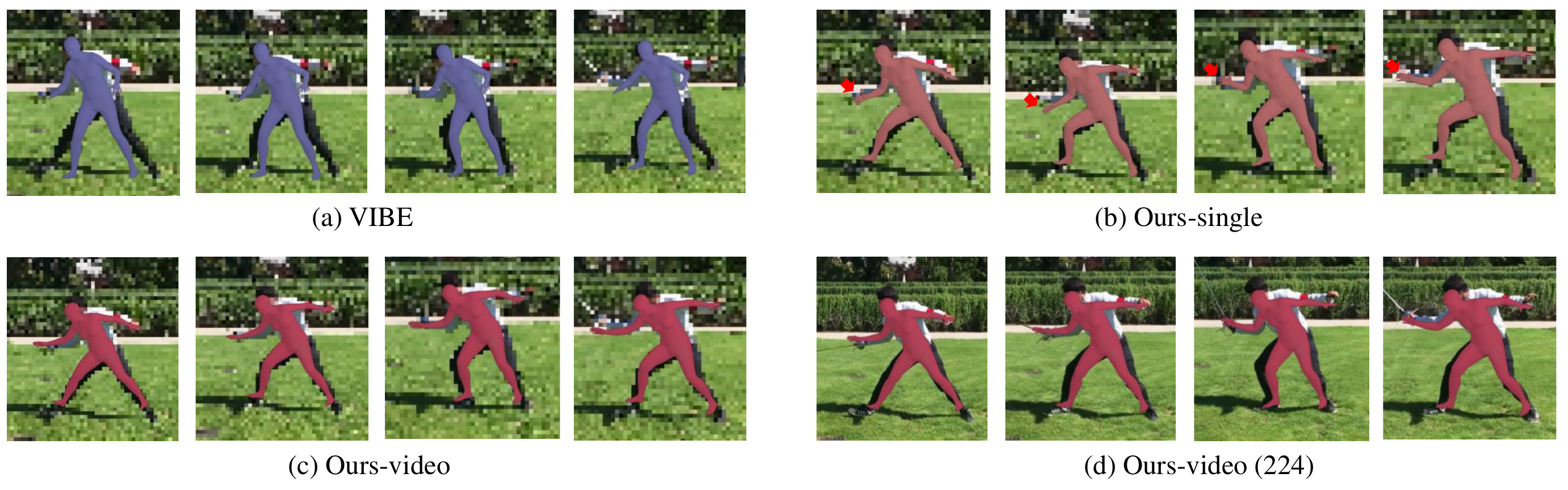}
	\vspace{-5mm}
	\caption{Qualitative evaluation of the proposed method on a low-resolution video sequence from 3DPW~\cite{3dpw}. Our video model (c) generates more accurate and temporally smoother predictions. The result of the corresponding high-resolution video (d) is also included as a reference. Note that the results of the single image model (b) is not as temporally-coherent (\eg the hand and body size).}
	\vspace{-3.5mm}
	\label{fig:vibe result}
\end{figure*}

\subsection{Implementation Details}
\textbf{Single-frame 3D human pose and shape.} We train our model and the baselines using a combination of 2D and 3D datasets similar to previous works \cite{kanazawa2018end,kolotouros2019spin}.
For the 3D datasets, we use Human3.6M \cite{human3.6} and MPI-INF-3DHP \cite{mpi-inf-3dhp} with ground truth of 3D keypoints, 2D keypoints, and SMPL parameters.
These datasets are mostly captured in constrained environments, and models trained on them do not generalize well to diverse images in real world.
For better performance on in-the-wild data, we also use 2D datasets including LSP~\cite{lsp}, LSP-Extended \cite{lsp-extend}, MPII \cite{mpii}, and MS COCO \cite{coco}, which only have 2D keypoint labels.
Similar to prior works~\cite{kanazawa2018end,kolotouros2019spin}, we assume the human bounding box is known, and the target human is located at the center of the bounding box.
We crop the human regions from the images and resize them to 224$\times$224.
Images with significant occlusions or small human are discarded from the dataset.
We consider human image resolutions ranging from 224 to 24.
As introduced in Section~\ref{sec:res-aware}, we split all the resolutions into $P=5$ ranges: $\{224, (224, 128], (128, 64], (64, 40], (40, 24]\}$, where the first range corresponds to the original high-resolution image $x_1$.
We obtain the lower-resolution images by downsampling the high-resolution images and resize them back to 224 with bicubic interpolation.
During training, we apply data augmentations to the images including Gaussian noise, color jitters, rotation, and random flipping.
For the loss functions, we set $\lambda_1=5$, $\lambda_2=5$, $\lambda_\text{s}=0.1$, and $\lambda_\text{f}=0.1$.
For contrastive learning, we set the size of the queue as $8192$ and $\tau=0.1$ in Eq.~\ref{eq:contrastive}.
Following \cite{chen2020simple}, the similarity function $s$ is equipped with a 2-layer MLP to better transform and measure the features.
As in \cite{kocabas2019vibe}, we initialize the baseline networks and our model with the parameters of \cite{kolotouros2019spin}.
We use the Adam algorithm~\cite{kingma2014adam} to optimize the network with a learning rate 5e-5.

Similar to \cite{kocabas2019vibe}, we conduct evaluations on a large in-the-wild dataset 3DPW~\cite{3dpw} with 3D joint ground truth to demonstrate the strength of our model in an in-the-wild setting.
We also provide results for constrained indoor images using the MPI-INF-3DHP~\cite{mpi-inf-3dhp} and H36M~\cite{human3.6} datasets.
Following \cite{kocabas2019vibe,kanazawa2018end,kolotouros2019spin}, we compute the procrustes aligned mean per joint position error (MPJPE-PA) and mean per joint position error (MPJPE) for measuring the 3D keypoint accuracy.
We mainly perform the evaluations on a challenging low resolution of 32$\times$32.
To analyze the robustness of our method on more diverse resolutions, we also compare the results of the resolutions 176, 96, 52, and 32 in Section~\ref{sec:robust}, which are the middle points of the divided training resolution ranges.

{\flushleft \textbf{Extension to low-resolution videos.} }
We use a 2-layer GRU and a 3-layer MLP as the temporal recurrent module.
We follow a similar paradigm as in VIBE \cite{kocabas2019vibe} to extract features before training the temporal recurrent module. The offline feature extraction allows larger batch size and faster training.
For the video datasets, we use the training set of InstaVariety~\cite{kanazawa2019learning}, MPI-INF- 3DHP~\cite{mpi-inf-3dhp}, 3DPW~\cite{3dpw}, and Human3.6M~\cite{human3.6}.
In addition, we use the large-scale 3D human motion dataset AMASS~\cite{amass2019} as the ground truth motion for adversarial training.
We refer to VIBE \cite{kocabas2019vibe} for more implementation details. %

{\flushleft \textbf{Texture reconstruction.}}
Each Conv block of our texture estimation network consists of two convolutional layers followed by Batch Normalization~\cite{bnorm} and ReLU activation~\cite{nair2010rectified}.
For the ReID loss, we use the model in \cite{sun2018beyond} as the pedestrian re-identification network for its simplicity and efficiency.
Similar to \cite{wang2019re}, we use 1401 person identities of the Market-1501 dataset~\cite{zheng2015scalable} for training, and the remaining 100 identities are used for testing.
We also use the SURREAL dataset~\cite{varol2017learning} to train the network such that the predicted texture can have a plausible-looking face~\cite{wang2019re}.
To synthesize the low-resolution data, we downsample the images to a height of 32 pixels which are further resized to 128 before fed into the network.
Due to the limitation of the SMPL model, the reconstructed human and the input image are not always well-aligned, and thus the pixel-wise evaluation metrics, such as PSNR and SSIM, cannot effectively measure the quality of the output image.
To solve this problem, we use the ReID loss (Eq.~\ref{eq:reid}) for evaluation which assesses the results from a semantic (identification) perspective and thus can better focus on the texture similarity of human bodies without affected by misalignment issues.
In addition, we also use the LPIPS~\cite{zhang2018unreasonable} to evaluate the perceptual quality of the reconstructed textures.

\begin{figure*}[t]
	\centering
	\includegraphics[width=0.93\textwidth]{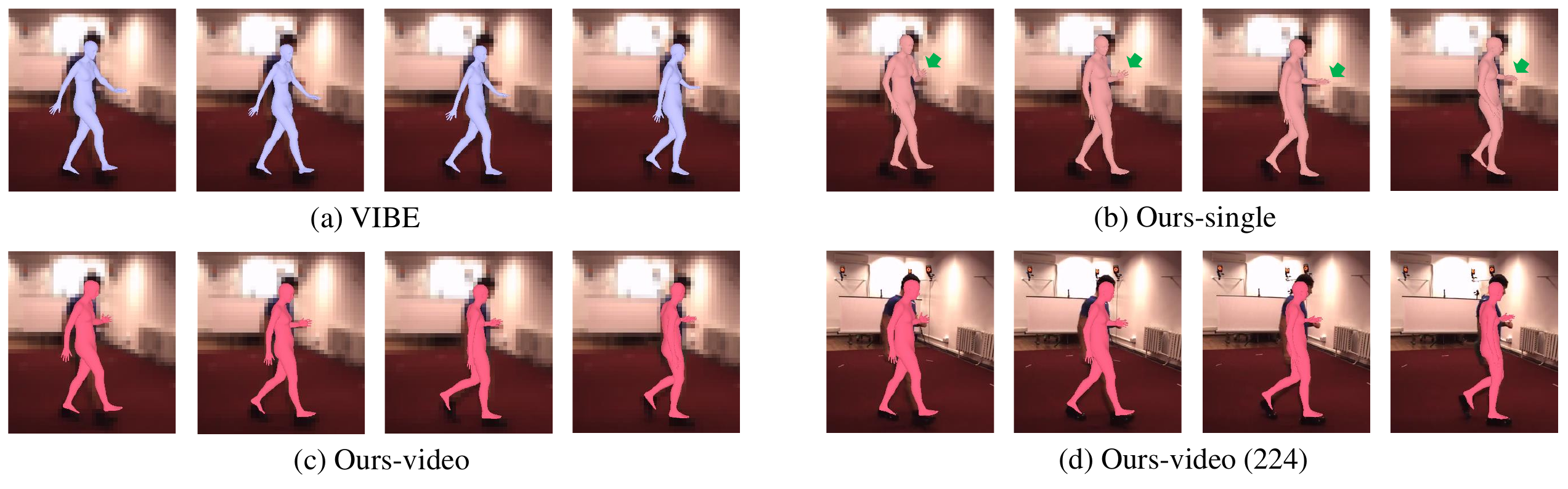}
	\vspace{-5mm}
	\caption{\revise{Qualitative evaluation of the proposed method on a low-resolution video sequence from H36M~\cite{human3.6}. Our video model (c) generates more accurate and temporally smoother predictions. The result of the corresponding high-resolution video (d) is also included as a reference. Note that the results of the single image model (b) is not as temporally-coherent (\eg the movement of the hand).}}
	\vspace{-3.5mm}
	\label{fig:vibe result2}
\end{figure*}

\subsection{Evaluation on Low-Resolution Single Image}
We compare against the state-of-the-art 3D human pose and shape estimation methods HMR~\cite{kanazawa2018end} and SPIN~\cite{kolotouros2019spin}, by fine-tuning them on different resolution images with the same training data and settings as our model.
In addition, since no previous approach has focused on the problem of low-resolution 3D human pose and shape estimation, we adapt the low-resolution image recognition algorithms to our task as new baselines, including both image super-resolution based~\cite{haris2018task} and feature enhancement based~\cite{tan2018feature}.
For the image super-resolution based method (denoted as ImgSR), we first use a state-of-the-art network RDN~\cite{SR-residual-dense} to super-resolve the low-resolution image, and the output is then fed into SPIN~\cite{kolotouros2019spin} for regressing the SMPL parameters.
Similar to \cite{haris2018task}, the network is trained to improve both the perceptual image quality and the 3D human pose and shape estimation accuracy.
For feature enhancement (denoted as FeaEN), we apply the strategy in \cite{tan2018feature} which uses a GAN loss to enhance the discriminative ability of the low-resolution features for better image retrieval performance.
Nevertheless, we find the WGAN~\cite{arjovsky2017wasserstein} used in the original work \cite{tan2018feature} does not work well in our experiments, and we instead use the LSGAN~\cite{mao2017least} combined with the basic loss (Eq.~\ref{eq:basic loss}) to train a stronger baseline network.

As shown in Table~\ref{tab:3dpw-mpi-h36m}, the proposed method generates consistently better results than the baseline approaches on the 3DPW, MPI-INF-3DHP, and H36M datasets for  challenging low-resolution (32$\times$32) images.
Note that we achieve significant improvement over the baselines on the 3DPW dataset, which demonstrates the effectiveness of the proposed method on the challenging in-the-wild images.
Similar to \cite{kolotouros2019spin,kanazawa2018end}, we also evaluate our approach on the auxiliary task of 2D human body segmentation on the test images of LSP. The proposed algorithm compares favorably against the baseline methods as shown in Table~\ref{tab:lsp}.
Furthermore, we provide a qualitative comparison against the baseline models in Figure \ref{fig:SOTA comparison}, where the proposed method generates higher-quality 3D human estimation results on the challenging low-resolution input.
We also provide visual examples of real low-resolution images obtained from the Internet in Figure~\ref{fig:teaser_real} and \ref{fig:real}, which further shows that the the proposed algorithm can well generalize to images in real world.

\subsection{Evaluation on Low-Resolution Videos}
In Section~\ref{sec:video}, we introduce a post-processing module to exploit the temporal information of videos.
To evaluate the performance of the proposed multi-frame network on low-resolution videos, we use the sequences of the 3DPW~\cite{3dpw} and the H36M~\cite{human3.6} datasets that are downsampled to 32$\times$32 pixels.
We follow VIBE~\cite{kocabas2019vibe} to pre-process the test data, where the severely-occluded frames are also included to ensure the continuity of the input frames.
For fair comparisons, we re-train the baseline network VIBE~\cite{kocabas2019vibe} on low-resolution videos with the same training data and training settings as our models.
As shown in Table~\ref{tab:video}, the proposed video model attains more accurate predictions than VIBE, which demonstrates that the RSC-Net can be well extended to handle low-resolution video input.

To measure the temporal consistency of the results, we also report the acceleration which represents the second-order derivative of the 3D joint motion, and the acceleration error that is calculated by the difference between ground-truth and predicted 3D acceleration of every joint.
As shown in Table~\ref{tab:video}, the proposed multi-frame model (``Ours-video'') can achieve better temporal coherence than the single-frame network (``Ours-single''), which shows the effectiveness of the temporal recurrent module. %
In addition, we provide qualitative comparisons in Figure~\ref{fig:vibe result} and \ref{fig:vibe result2}, where our video model generates more accurate and temporally smoother results.

\begin{table}[t]
	\centering
	\caption{\label{tab:lsp} Results on the LSP dataset~\cite{lsp}. ``H-acc'' and ``H-f1'' denote the accuracy and f1 score of the human silhouettes, while ``P-acc'' and ``P-f1'' denote those of body part segmentation.}
	\begin{tabular}{l*{4}{>{\centering\arraybackslash}p{0.06\textwidth}}}
		\toprule
		Methods~ & H-acc          & H-f1            & P-acc          & P-f1            \\
		\midrule
		HMR                      & 89.33          & 0.8273          & 84.73          & 0.5145          \\
		SPIN                     & 90.00          & 0.8384          & 87.09          & 0.5998          \\
		ImgSR                     & 89.75          & 0.8308          & 87.07          & 0.5955          \\
		FeaEN                    & 89.57          & 0.8287          & 86.71          & 0.5843          \\
		Ours                     & \textbf{90.18} & \textbf{0.8412} & \textbf{87.43} & \textbf{0.6159} \\
		\bottomrule
	\end{tabular}
\end{table}

\begin{table}[t]
	\centering
	\caption{\label{tab:video} Quantitative evaluation on low-resolution videos.
		``ACC'' and ``ACC-ERR'' represent the acceleration and the acceleration error, respectively.
	}
	\vspace{-2mm}
	\begin{tabular}{clcccc}
		\toprule
		& Methods	& MPJPE &	MPJPE-PA &	ACC & ACC-ERR \\ \midrule
		\multirow{3}{*}{3DPW} & VIBE~\cite{kocabas2019vibe} & 135.05 & 77.94 &  52.99 & 54.07 \\
		& Ours-single & 141.87 & 75.04 &  90.61 & 91.12 \\
		& Ours-video & \bf 123.14 & \bf 72.64 & \bf 48.63 & \bf	49.58 \\		\midrule
    \revises \multirow{3}{*}{H36M} & \revises VIBE~\cite{kocabas2019vibe} & \revises 122.16 & \revises 82.19 & \revises 72.67 & \revises 68.48 \\
    & \revises Ours-single & \revises 135.09 & \revises 75.72 &\revises  170.35 & \revises 162.33 \\
    & \revises Ours-video & \revises \bf 84.10 &\revises  \bf 60.04 & \revises \bf 63.23 &\revises  \bf 60.80 \\ \bottomrule

	\end{tabular}
\end{table}

\begin{table}[t]
	\centering
	\caption{\label{tab:texture}
		Quantitative evaluation of the proposed texture estimation network against the baseline model~\cite{wang2019re} on the test set of Market-1501~\cite{zheng2015scalable}.
		The original ``Baseline'' uses the HMR~\cite{kanazawa2018end} for 3D body mesh prediction and the OpenDR~\cite{loper2014opendr} for differentiable rendering.
		We re-train the baseline with the 3D body mesh predicted by our RSC-Net, which is denoted as ``Baseline$^\dag$''.
		We further replace the OpenDR~\cite{loper2014opendr} with the Pytorch3D~\cite{ravi2020pytorch3d}, leading to an even stronger baseline, \ie ``Baseline$^*$''.
	}
	\vspace{-2mm}
	\begin{tabular}{lcccc}
		\toprule
		Metrics	& Baseline & Baseline$^\dag$ & Baseline$^*$ & TexGlo \\
		\midrule
		ReID loss & 60.61 & 60.41 & 60.21 & \bf 59.89 \\
\revises LPIPS & \revises 0.2170 & \revises 0.2156 & \revises 0.2120 & \revises \bf 0.2094 \\
		\bottomrule
	\end{tabular}
	\vspace{-2mm}
\end{table}

\begin{figure}[t]
	\centering
	\includegraphics[width=0.48\textwidth]{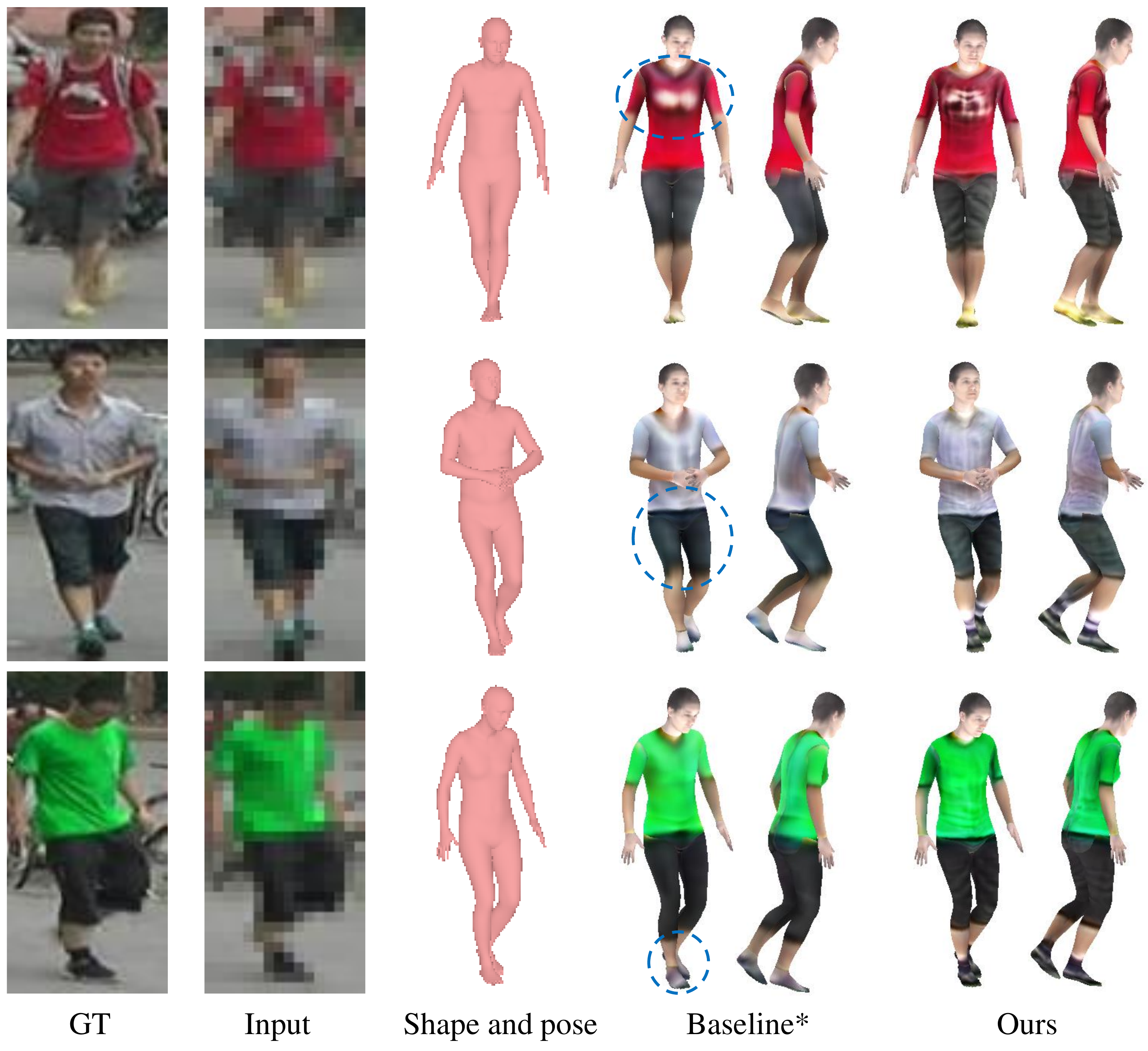}
	\vspace{-2mm}
	\caption{Qualitative evaluation of the texture estimation network. HR represents the corresponding high-resolution image.}
	\vspace{-1.5mm}
	\label{fig:tex result}
\end{figure}

\begin{figure*}[t]
	\centering
	\begin{tabular}{cccc}
		\hspace{-3mm} \includegraphics[width=0.24\textwidth]{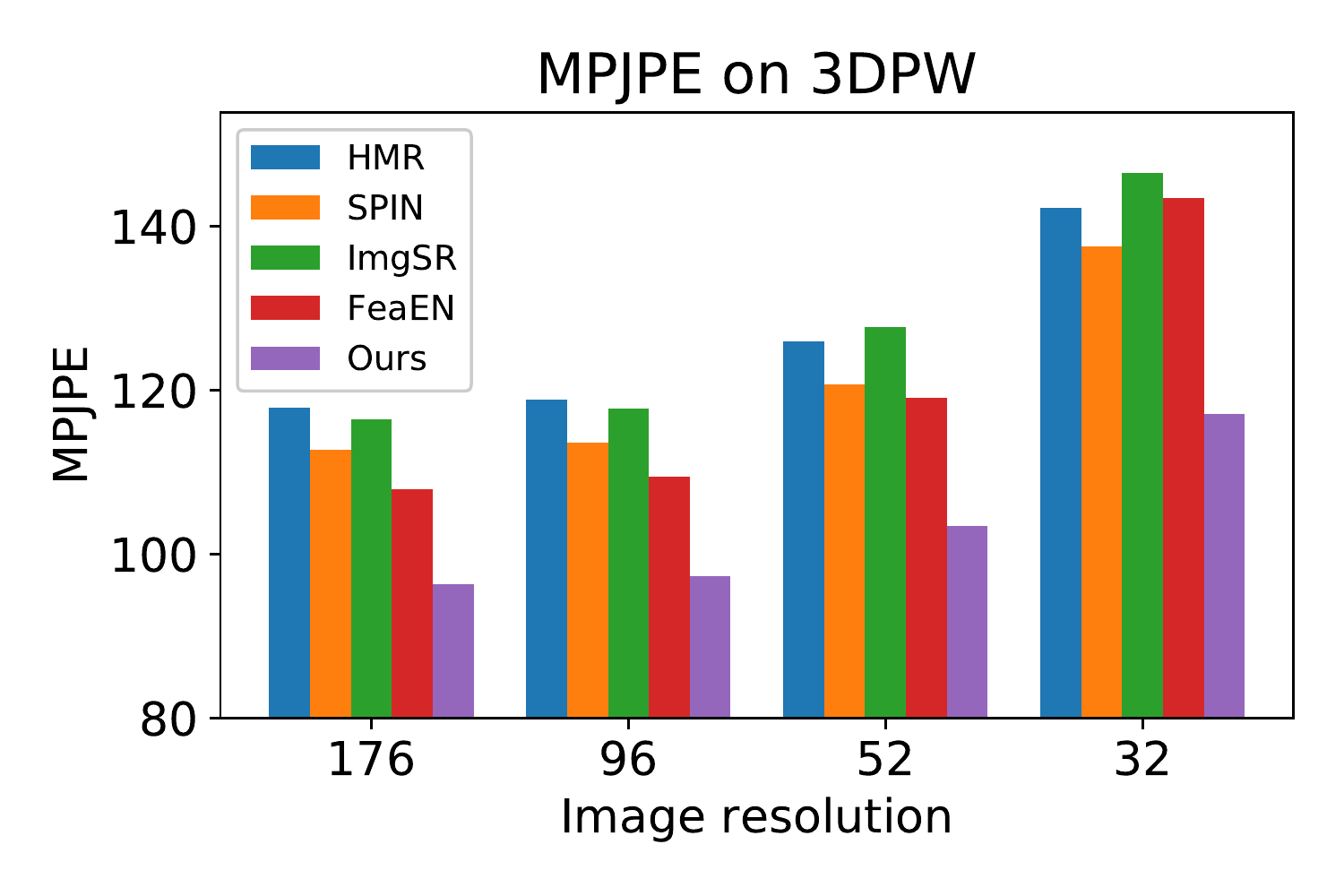}  & \hspace{-5mm} \includegraphics[width=0.24\textwidth]{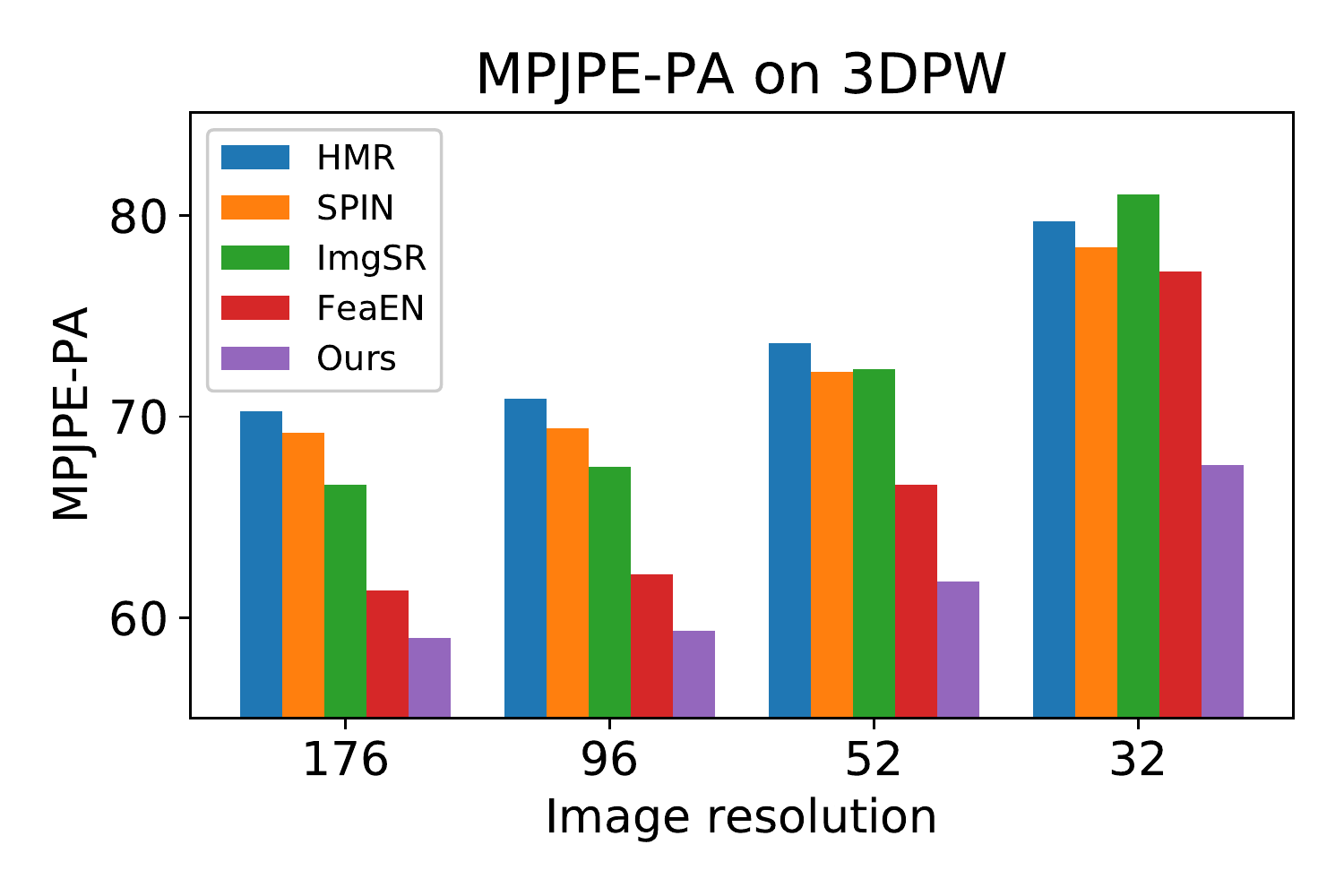}  &
		\hspace{-1mm} \includegraphics[width=0.24\textwidth]{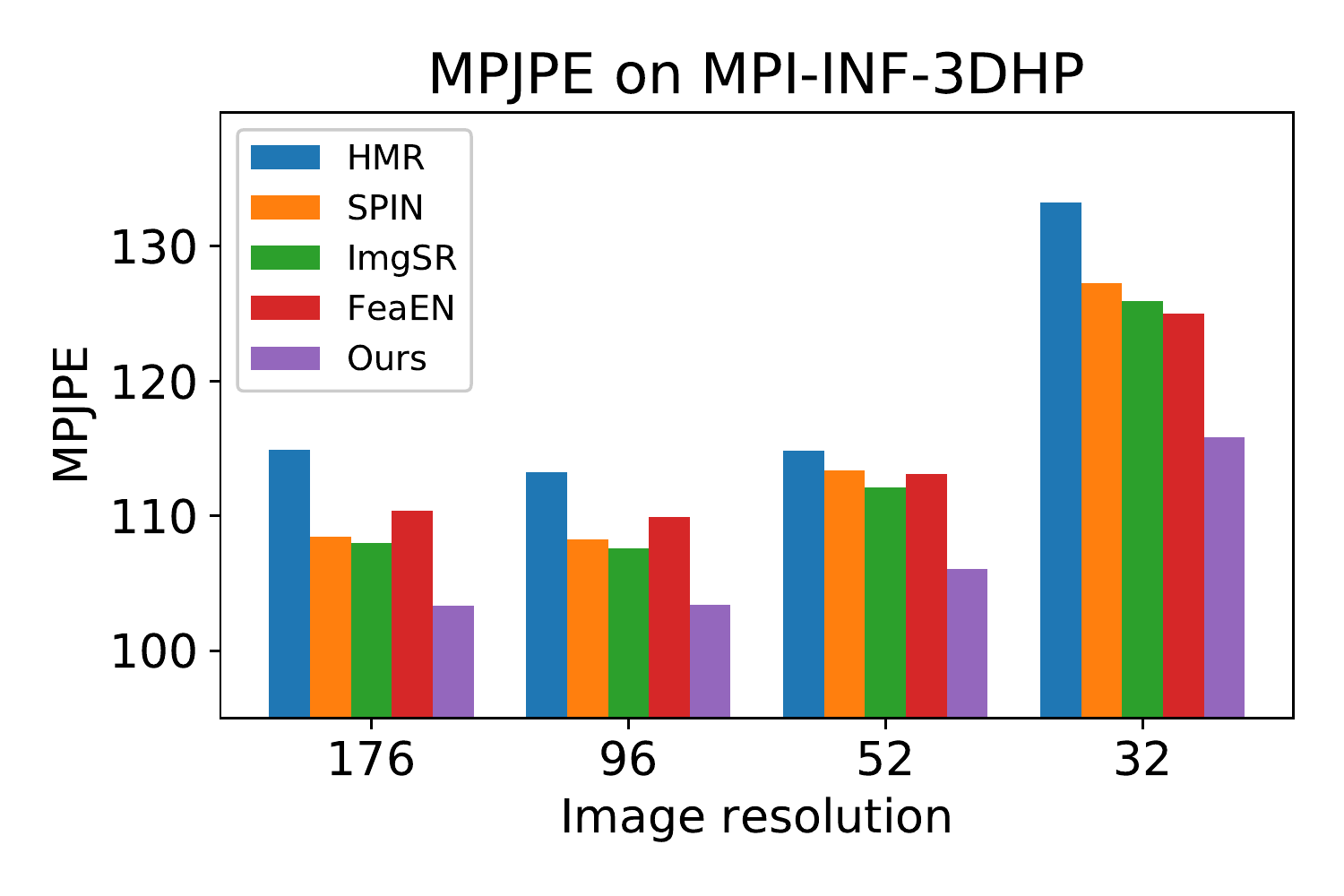}  & \hspace{-5mm} \includegraphics[width=0.24\textwidth]{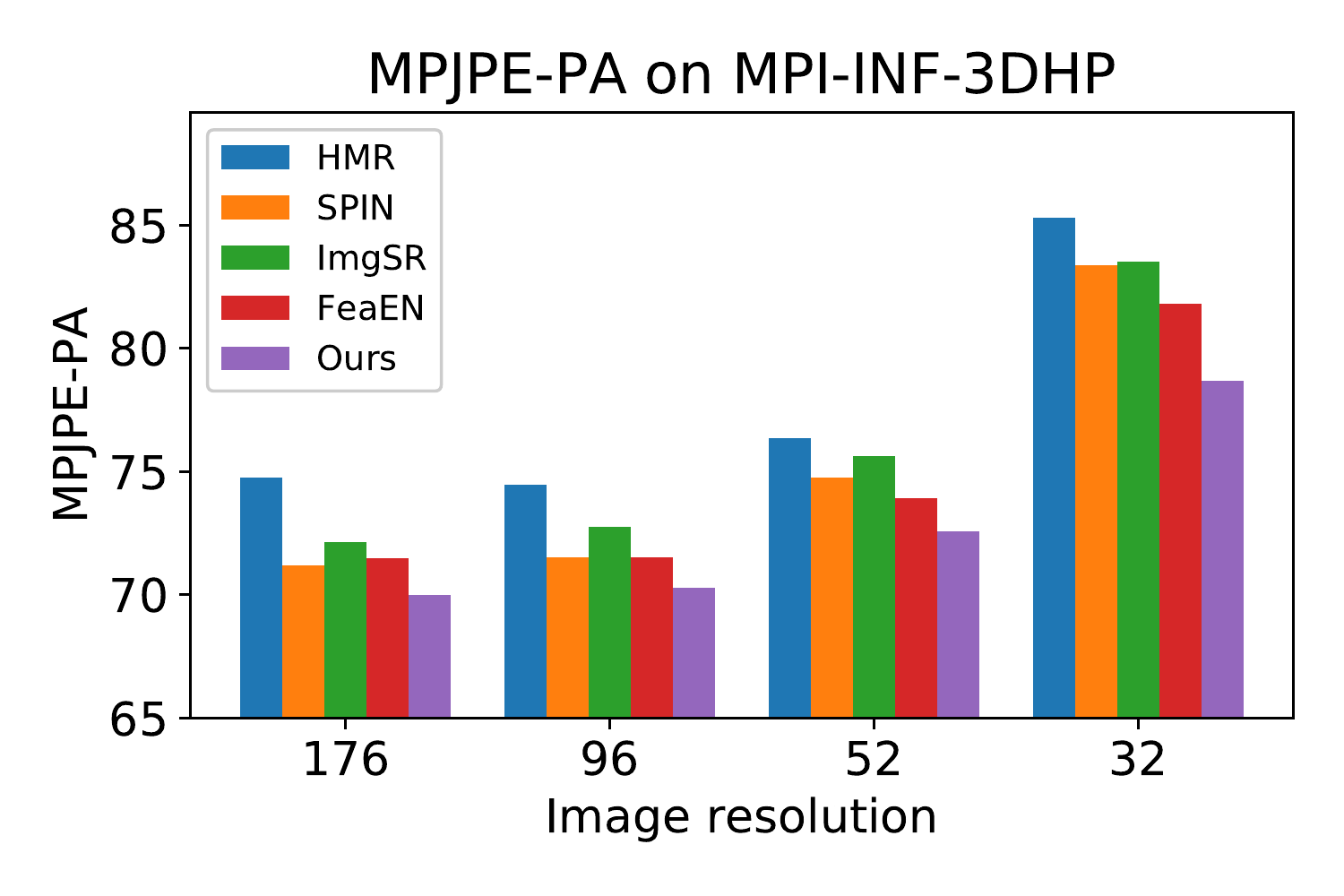}  \\
	\end{tabular}
	\vspace{-3.5mm}
	\caption{Robustness of the RSC-Net on different low-resolution images.}
	\label{fig:bar_3dpw}
\end{figure*}

\begin{figure*}[t]
	\centering
	\includegraphics[width=0.95\textwidth]{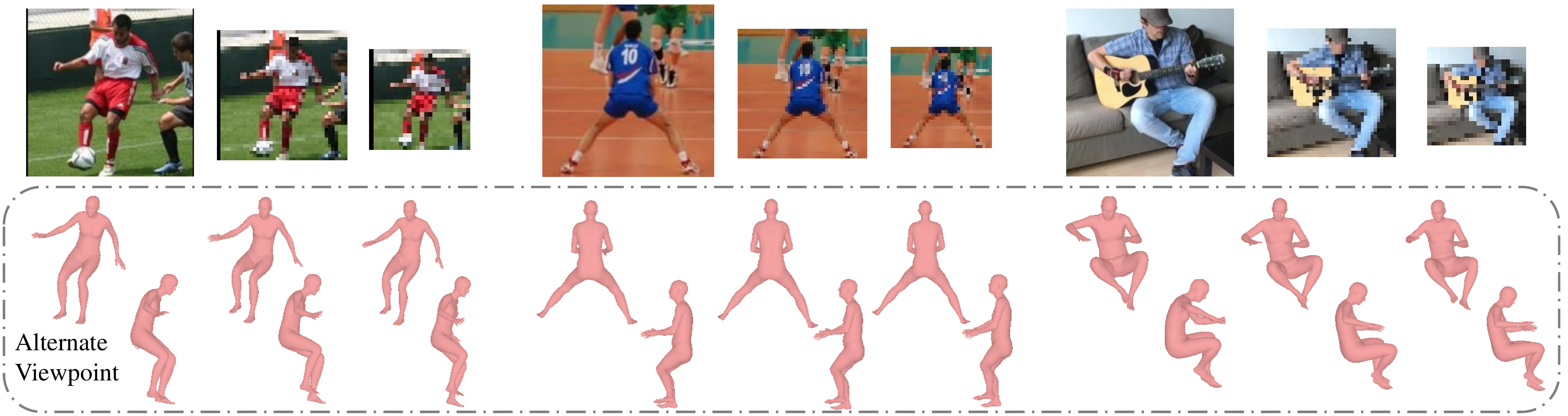}
	\vspace{-2mm}
	\caption{The proposed method can achieve robust 3D human pose and shape estimation for different resolution inputs. The resolutions for the images are 224, 52, and 32.}
	\label{fig:different resolution}
\end{figure*}

\subsection{Evaluation of the Texture Estimation Network}
In Table~\ref{tab:texture}, we compare our method with the baseline model~\cite{wang2019re} for low-resolution human texture reconstruction.
Our texture estimation network with the global module is named as TexGlo.
The original ``Baseline'' uses the HMR~\cite{kanazawa2018end} for 3D body mesh prediction and the OpenDR~\cite{loper2014opendr} for differentiable rendering, whereas we use the proposed \netname{} and the Pytorch3D renderer~\cite{ravi2020pytorch3d}.
For better comparison, we re-train the baseline with the 3D body mesh predicted by our network (``Baseline$^\dag$'').
We further replace the OpenDR with the Pytorch3D, which leads to an even stronger baseline, \ie ``Baseline$^*$''.
As shown in Table~\ref{tab:texture}, our method can achieve better results (lower ReID errors and LPIPS) than all the baselines, which demonstrates the effectiveness of the proposed texture estimation network.

\revise{For more comprehensive analysis, we further conduct a user study to evaluate the texture results. 
This study uses 10 images randomly selected from the test set of Market-1501~\cite{zheng2015scalable}, and 21 subjects are asked to compare the textures reconstructed by ``Baseline$^*$'' and our algorithm. 
92.4\% of the answers prefer the textures generated by the proposed network, which shows the effectiveness of our method.
In addition, we also present visual examples in Figure~\ref{fig:tex result}, where the proposed method recovers more vivid colors and finer details.}

\begin{table}[t]
	\centering
	\vspace{0mm}
	\caption{\label{tab:high-res}
	Comparison of the MPJPE on high-resolution images.
	``SPIN-o'' represents the original SPIN~\cite{kolotouros2019spin} model specifically trained for high-resolution images, whereas ``SPIN'' is trained on all the different resolution images. Numbers in bold indicate the best in each column, and underscored numbers indicate the second.}
	\vspace{-2mm}
	\begin{tabular}{l*{1}{>{\centering\arraybackslash}p{0.05\textwidth}}  *{1}{>{\centering\arraybackslash}p{0.05\textwidth}} >{\centering\arraybackslash}p{0.002\textwidth} *{1}{>{\centering\arraybackslash}p{0.05\textwidth}}  *{1}{>{\centering\arraybackslash}p{0.05\textwidth}}}
		\toprule
		\multirow{3}{*}{Methods~~} & \multicolumn{2}{c}{3DPW}           &               & \multicolumn{2}{c}{MPI-INF-3DHP}                          \\
		\cmidrule{2-3} \cmidrule{5-6}
		& \multicolumn{1}{c}{224} & \multicolumn{1}{c}{32} & & \multicolumn{1}{c}{224} & \multicolumn{1}{c}{32} \\
		\midrule
		SPIN-o            & \underline{97.18}            &    149.31           && \bf {103.39}    & 147.67             \\
		SPIN & 113.24 & \underline{137.61} &&  108.80  & \underline{127.27} \\
		Ours & \bf 96.60  &  \bf 117.12 &&  \underline{103.50}  & \bf 115.80  \\
		\bottomrule
	\end{tabular}
	\vspace{-2mm}
\end{table}

\section{Analysis and Discussion} \label{sec:analysis}
In this section, we first analyze the robustness of the proposed RSC-Net for different resolution images.
We then present a comprehensive ablation study to demonstrate the effectiveness of the resolution-aware network, the self-supervision loss, the contrastive learning loss.
Finally, we provide more analysis of our texture estimation network.

\subsection{Robustness to Different Resolutions} \label{sec:robust}
As introduced in Section~\ref{sec:SS} and \ref{sec:contra}, the RSC-Net is trained by enforcing the scale consistency of the predicted results and features. As a result, the proposed algorithm is able to achieve robust performance on different resolution images as shown in Figure~\ref{fig:bar_3dpw}, where our method achieves consistently better results than the state-of-the-art approaches.

In addition, we also conduct evaluations on high-resolution images to demonstrate the robustness of the proposed method.
As shown in Table~\ref{tab:high-res}, our RSC-Net is able to compete with the state-of-the-art high-resolution model~\cite{kolotouros2019spin} (``SPIN-o'' in Table~\ref{tab:high-res}) on high-resolution input.
Further, the proposed method significantly outperforms ``SPIN-o'' on low-resolution images, as ``SPIN-o'' has only been trained for high-resolution images whereas our network is trained for all the different resolution images.
On the other hand, while fine-tuning ``SPIN-o'' on more diverse resolutions (\ie ``SPIN'' in Table~\ref{tab:high-res}) can achieve better results on low-resolution input, the performance of SPIN suffers from a significant drop on high-resolution images.
In other words, SPIN needs to sacrifice its performance on high-resolution images to achieve good results on other resolutions, which further demonstrates the robustness of the proposed algorithm.
In addition, we present visual examples in Figure~\ref{fig:different resolution} where our method generates high-quality 3D body pose and shape for different resolutions.

\begin{table*}[t]
	\centering
	\footnotesize
	\caption{\label{tab:ablation} Ablation study of the proposed method. Ba: baseline network with basic loss function, RA: resolution-aware network with basic loss function, SS: self-supervision loss, MS: MSE feature loss, CD: cosine distance feature loss, CL: contrastive learning feature loss. %
	}
	\vspace{-2mm}
	\begin{tabular}{l*{4}{>{\centering\arraybackslash}p{0.06\textwidth}} >{\centering\arraybackslash}p{0.005\textwidth} *{4}{>{\centering\arraybackslash}p{0.06\textwidth}}}
		\toprule
		\multirow{2}{*}{Methods} & \multicolumn{4}{c}{MPJPE} && \multicolumn{4}{c}{MPJPE-PA} \\
		\cmidrule{2-5}  \cmidrule{7-10}
		&  176  & 96  & 52  & 32 & &  176  & 96  & 52  & 32  \\
		\midrule
		Ba                 &    112.26 & 115.18 &  124.88  & 143.63 &  & 65.04  & 66.41 &   71.12  &  79.43   \\
		Ba+SS           &  107.51 & 109.58  & 116.54     &  128.88   &  &   62.32 & 63.27  &  66.78 & 72.49   \\
		Ba+SS+CL  & 106.06 & 107.75 & 112.44 & 123.63 & & 63.25 & 64.07 & 66.04 & 70.86 \\
		RA                &    111.55  &  112.18   &  118.70   & 135.29  & &   64.53   & 68.88  &  68.01 &  75.49    \\
		RA+SS        &   102.56 & 104.18 & 110.17     & 124.23   &  &  60.17  &  60.84  & 63.71  & 69.87  \\
		RA+SS+MS          &  105.96 & 106.15 & 111.33 & 124.85  & &    60.90 &  61.76 &  64.55 & 70.40  \\
		RA+SS+CD        &   104.95 & 105.96  & 111.41 &  125.08 & &  61.29 & 61.91 & 64.30 & 70.17 \\
		RA+SS+CL~         &  \bf 96.36   &  \bf 97.36    & \bf 103.49   &  \bf 117.12  &   & \bf 58.98 & \bf 59.34   & \bf 61.81 & \bf 67.59 \\
		\bottomrule
	\end{tabular}
\end{table*}

\begin{table*}[t]
	\centering
	\caption{\label{tab:analysis} Analysis of the alternative training strategies. PT: Progressive Training, SS-o: original self-supervision loss, SS-h: only using the highest-resolution for supervision.}
	\vspace{-2mm}
	\begin{tabular}{l*{4}{>{\centering\arraybackslash}p{0.06\textwidth}} >{\centering\arraybackslash}p{0.005\textwidth} *{4}{>{\centering\arraybackslash}p{0.06\textwidth}}}
		\toprule
		\multirow{2}{*}{Methods~~} & \multicolumn{4}{c}{MPJPE} & & \multicolumn{4}{c}{MPJPE-PA} \\
		\cmidrule{2-5}  \cmidrule{7-10}
		&  176  & 96  & 52  & 32 &  &  176  & 96  & 52  & 32  \\
		\midrule
		w/o PT                &     105.11     &  106.60   & 113.41  & 127.05 &  &    61.46    &  62.22   &  65.47   & 71.30    \\
		w/ SS-o           & 143.31  & 142.32  &  145.61 &  156.25  &  &  77.75 & 77.51 & 79.06  &  82.97  \\
		w/ SS-h            &  104.16     & 105.24  &  109.94    &  122.01   &  &     62.46  &  62.73   &  64.47   & 68.89 \\
		full model       &   \bf  96.36   &  \bf 97.36    & \bf 103.49   &  \bf 117.12  &   & \bf 58.98 & \bf 59.34   & \bf 61.81 & \bf 67.59 \\
		\bottomrule
	\end{tabular}
\end{table*}

\begin{figure}[t]
	\centering
	\includegraphics[width=0.49\textwidth]{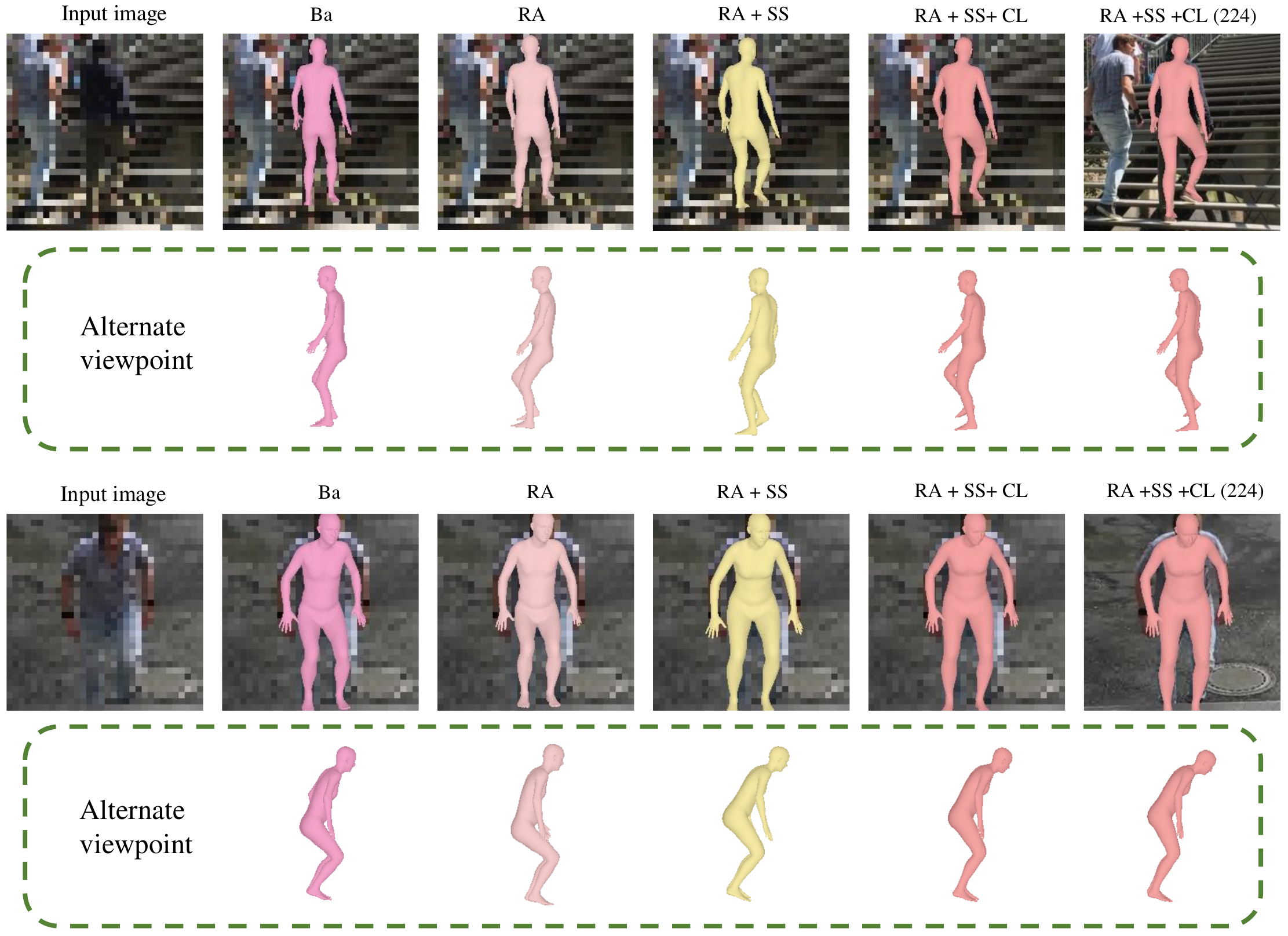}
	\vspace{-3mm}
	\caption{Visual examples which show the effectiveness of the three components of the RSC-Net, \ie the resolution-aware network, the self-supervision loss, and the contrastive learning feature loss.}
	\vspace{-2.5mm}
	\label{fig:visual_ablation}
\end{figure}

\subsection{Ablation Study}
In Table~\ref{tab:ablation}, we provide an ablation study using the 3DPW dataset to evaluate the proposed resolution-aware network, the self-supervision loss, and the contrastive feature loss.
We first compare the proposed resolution-aware network with the baseline model ResNet-50~\cite{kanazawa2018end,resnet}.
As shown by ``RA'' (resolution-aware) and ``Ba'' (baseline) in Table \ref{tab:ablation}, our network can obtain slightly better results than the baseline network with the basic loss (Eq.~\ref{eq:basic loss}) as loss function.
Further, we can achieve a more significant improvement over the baseline when adding the self-supervision loss (Eq.~\ref{eq:self-supervision-v2}) and the contrastive feature loss (Eq.~\ref{eq:feature loss}) for training, \ie ``RA+SS'' \vs ``Ba+SS'' and ``RA+SS+CL'' \vs ``Ba+SS+CL'' in Table \ref{tab:ablation}, which further demonstrates the effectiveness of the resolution-aware structure.

Second, we use the self-supervision loss in~Eq.~\ref{eq:self-supervision-v2} to exploit the consistency of the outputs of the same input image with different resolutions.
By comparing ``RA+SS'' against ``RA'' in Table \ref{tab:ablation}, we show that the self-supervision loss is important for addressing the weak supervision issue of  3D human pose and shape estimation and thus effectively improves the results.
The comparison between ``Ba+SS'' and ``Ba'' also leads to similar conclusions.

In addition, we propose to enforce the consistency of the features across different image resolutions.
However, a normally-used MSE loss does not work well as shown in ``RA+SS+MS'' of Table \ref{tab:ablation}, which is mainly due to that the unimodal losses are not effective in modeling the correlations between high-dimensional vectors and can be easily affected by noise and insignificant structures in the embedded features~\cite{oord2018representation}.
In contrast, the proposed contrastive feature loss can more effectively improve the feature representations by maximizing the mutual information across the features of different resolutions, and achieve better results as in ``RA+SS+CL'' of Table \ref{tab:ablation}.
Note that we adopt the cosine similarity function in the contrastive feature loss (Eq.~\ref{eq:contrastive}) similar to prior methods~\cite{oord2018representation,he2019momentum,tian2019contrastive}.
Alternatively, one may only use the cosine distance for measuring the distance of two features instead of using the whole contrastive loss (Eq.~\ref{eq:contrastive}).
Nevertheless, this strategy does not work well as shown by ``RA+SS+CD'' in Table \ref{tab:ablation}, which demonstrates the effectiveness of the proposed contrastive learning scheme.

Furthermore, we present several visual examples in Figure~\ref{fig:visual_ablation}, which intuitively illustrate that all the three components of the RSC-Net (RA, SS, and CL) essentially improve the estimated 3D human pose and shape.

\begin{table}[t]
	\centering
	\caption{\label{tab:texture2} Effectiveness of the proposed texture estimation network. ``Up'' represents the extra upsampling layer. ``Glo'' is the global context module.  ``Glo-C'' and ``Glo-S'' represent fusing the global features with concatenation and point-wise summation, respectively.}
	\vspace{-2mm}
	\begin{tabular}{lcccc}
		\toprule
		Methods	& w/o Up \& Glo  & w/o Glo & w/ Glo-C & w/ Glo-S \\
		\midrule
		ReID loss & 60.21 & 60.14 & 60.25 & \bf 59.89 \\
		\bottomrule
	\end{tabular}
\end{table}

\begin{figure}[t]
	\centering
	\includegraphics[width=0.49\textwidth]{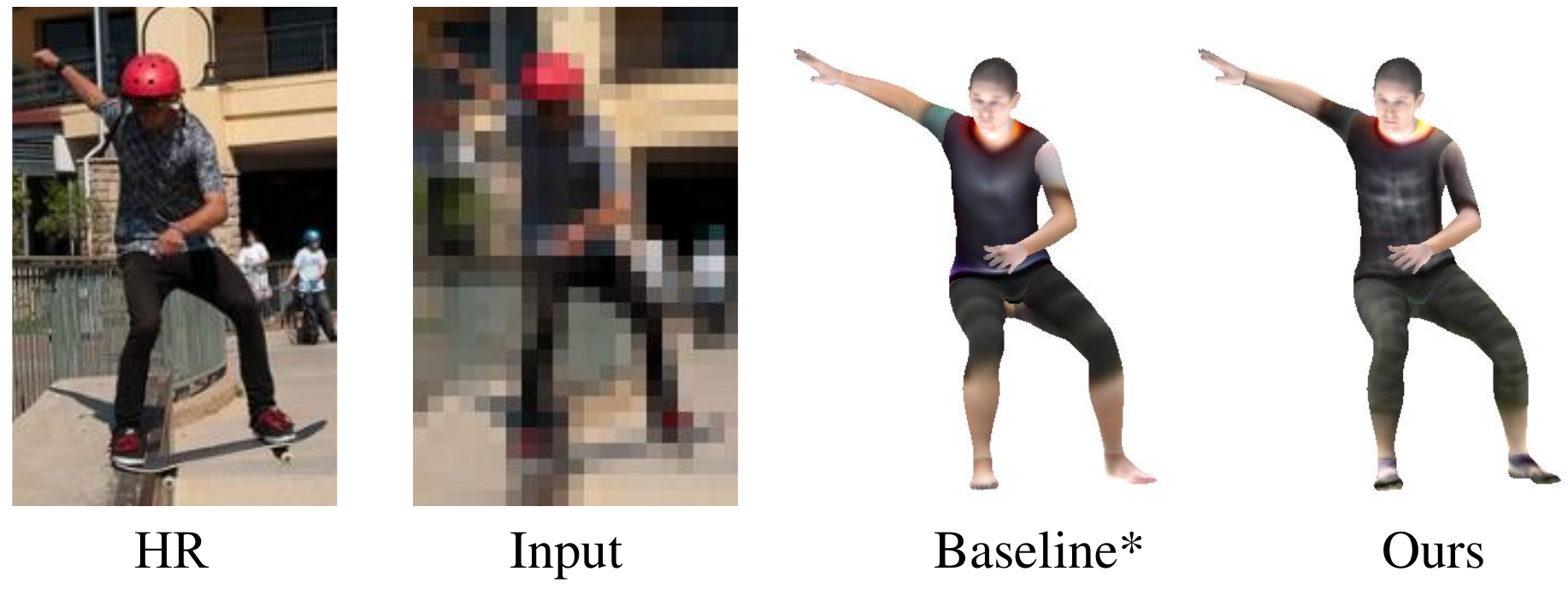}
	\vspace{-5mm}
	\caption{Limitation of the texture estimation network. The proposed method is not able to recover the textures well as the image has a rare color around the human head.}
	\vspace{-2.5mm}
	\label{fig:limit}
\end{figure}

\vspace{1mm}
\noindent \textbf{Analysis of training strategies.}
We also provide a detailed analysis of the alternative training strategies of our model.
First, as described in Section~\ref{sec:res-aware}, we train our model as well as the baselines in a progressive manner to deal with the challenging multi-resolution input.
As shown in the first row of Table \ref{tab:analysis} (``w/o PT''), directly training the model for all image resolutions without the progressive strategy leads to degraded results.

Second, the original self-supervision loss (Eq.~\ref{eq:self-supervision-v1}) treats the images under different augmentations equally, while we are generally more confident in the high-resolution predictions.
Therefore, we propose a directional self-supervision loss in Eq.~\ref{eq:self-supervision-v2} to exploit this prior knowledge.
As shown in the second row of Table \ref{tab:analysis} (``w/ SS-o''), using the original self-supervision loss (Eq.~\ref{eq:self-supervision-v1}) is not able to achieve high-quality results, as the network can minimize Eq.~\ref{eq:self-supervision-v1} by simply degrading the high-resolution predictions without improving the low-resolution ones (note the significant performance gap between ``w/ SS-o'' and ``full model'').
In addition, we provide hierarchical supervision for low-resolution images in Eq.~\ref{eq:self-supervision-v2} which can act as soft targets during training.
As shown in Table \ref{tab:analysis}, only using the highest-resolution predictions as guidance (``w/ SS-h'') cannot produce as good results as the proposed approach (``full model'').

\subsection{More Analysis of the Texture Estimation Network}
As introduced in Section~\ref{sec:texture}, the main difference between the baseline~\cite{wang2019re} and the proposed texture estimation network is the global context module which can capture the information of the human body globally and thereby generate higher-quality UV maps.
As shown in Table~\ref{tab:texture2}, the network without the global context module (``w/o Glo'') does not perform as well as our method (``w/ Glo-S'').
In addition, we use an extra upsampling layer before the output layer of the encoder-decoder architecture (Figure~\ref{fig:tex_overview}(b)), which allows the network to generate more details in the predicted UV map.
The effectiveness of this design has also been shown in Table~\ref{tab:texture2} (``w/o Up \& Glo'').
Note that we use the point-wise summation to fuse the extracted global context information back to the input feature maps in Figure~\ref{fig:tex_overview}(c), as we empirically find that the normally-used concatenation layer (``w/ Glo-C'') does lead to satisfying results.

\vspace{1mm}
{\noindent \textbf{Limitation.}}
Although the proposed algorithm is able to predict high-quality texture maps for low-resolution humans, the training process of this model is based on the ReID loss in Eq.~\ref{eq:reid}, which essentially relies on a pre-trained pedestrian re-identification network.
Thus, it has the same limitations as the state-of-the-art high-resolution texture estimation method~\cite{wang2019re} and is likely to fail when the input is not a common pedestrian scene.
As shown in Figure~\ref{fig:limit}, the proposed method is not able to recover the textures well as the image has a rare color around the human head. Our future work will focus on developing a texture estimation network that can better generalize to more diverse images.

\section{Conclusion}
In this work, we have studied the challenging problem of predicting 3D human pose and shape from low-resolution images and presented an effective solution for it.
We propose a resolution-aware neural network which can deal with different resolution images using a single model.
For training the proposed network, we propose a directional self-supervision loss which can exploit the output consistency across different resolutions to remedy the issue of lacking accurate 3D labels.
In addition, we introduce a contrastive feature loss which is more effective than MSE for measuring high-dimensional feature vectors and helps learn better feature representations.
Our method performs favorably against the state-of-the-art methods on different resolution images and achieves high-quality results for low-resolution 3D human pose and shape estimation.
Further, while we focus on learning 3D human pose and shape in this work, the proposed methodology is generic, and we expect its application in other low-resolution recognition problems.

\ifCLASSOPTIONcaptionsoff
  \newpage
\fi

\bibliographystyle{IEEEtran}
\bibliography{egbib}

\end{document}